\documentclass[journal]{IEEEtran}

\usepackage{amsmath}
\usepackage{multirow}
\usepackage{graphicx}

\usepackage{times}
\usepackage{epsfig}
\usepackage{graphicx}
\usepackage{amsmath}
\usepackage{amssymb}

\usepackage{comment}
\usepackage{color}

\usepackage{pifont}
\usepackage[dvipsnames,svgnames,x11names]{xcolor}
\definecolor{citecolor}{RGB}{119,185,0} 
\usepackage[pagebackref=false,breaklinks=true,letterpaper=true,colorlinks,citecolor=citecolor,bookmarks=false]{hyperref}

\usepackage{algorithm, algpseudocode}

\def\eg{\emph{e.g.}} 
\def\ie{\emph{i.e.}} 
\def\etal{\emph{et~al.}} 

\makeatletter
\newlength\savewidth\newcommand\shline{\noalign{\global\savewidth\arrayrulewidth
  \global\arrayrulewidth 1pt}\hline\noalign{\global\arrayrulewidth\savewidth}}
\def\endthebibliography{%
 \def\@noitemerr{\@latex@warning{Empty `thebibliography' environment}}%
 \endlist
}
\begin{document}
\pagestyle{plain}
%
\title{Each Part Matters: Local Patterns Facilitate \\ Cross-view Geo-localization}
%
%
%

\author{Tingyu~Wang,
        Zhedong~Zheng,
        Chenggang Yan,
        Jiyong Zhang,~\IEEEmembership{Member,~IEEE},
        Yaoqi Sun,
        Bolun Zheng,
        and~Yi~Yang,~\IEEEmembership{Senior~Member,~IEEE}
\thanks{Copyright~\copyright~2021 IEEE. Personal use of this material is permitted. However, permission to use this material for any other purposes must be obtained from the IEEE by sending an email to pubs-permissions@ieee.org.}
\thanks{Tingyu Wang, Chenggang Yan, Jiyong Zhang, Yaoqi Sun and Bolun Zheng are with the Intelligent Information Processing Lab, Hangzhou Dianzi University, Hangzhou 310018, China (e-mail: wongtyu@hdu.edu.cn; cgyan@hdu.edu.cn; jzhang@hdu.edu.cn; syq@hdu.edu.cn; blzheng@hdu.edu.cn). Chenggang Yan is the Corresponding Author.}
\thanks{Zhedong Zheng, Yi Yang are with the Australian Artificial Intelligence Institute, University of Technology Sydney, NSW 2007, Australia (e-mail: Zhedong.Zheng@student.uts.edu.au; Yi.Yang@uts.edu.au).}
\thanks{This work was partially sponsored by Zhejiang Lab’s International Talent Fund for Young Professionals (No.ZJ2020GZ021).}
}

\markboth{Journal of \LaTeX\ Class Files,~Vol.~14, No.~8, August~2015}%
{Shell \MakeLowercase{\textit{et al.}}: Bare Demo of IEEEtran.cls for IEEE Journals}
%



\newcommand{\zznote}[1]{\textcolor{blue}{ZZ:#1}}
\newcommand{\wtynote}[1]{\textcolor{red}{WTY:#1}}
\maketitle

\begin{abstract}
Cross-view geo-localization is to spot images of the same geographic target from different platforms, \eg, drone-view cameras and satellites. It is challenging in the large visual appearance changes caused by extreme viewpoint variations. Existing methods usually concentrate on mining the fine-grained feature of the geographic target in the image center, but underestimate the contextual information in neighbor areas. In this work, we argue that neighbor areas can be leveraged as auxiliary information, enriching discriminative clues for geo-localization. Specifically, we introduce a simple and effective deep neural network, called Local Pattern Network (\textbf{LPN}), to take advantage of contextual information in an end-to-end manner. 
Without using extra part estimators, LPN adopts a square-ring feature partition strategy, which provides the attention according to the distance to the image center. It eases the part matching and enables the part-wise representation learning. Owing to the square-ring partition design, the proposed LPN has good scalability to rotation variations and achieves competitive results on three prevailing benchmarks, \ie, University-1652, CVUSA and CVACT. Besides, we also show the proposed LPN can be easily embedded into other frameworks to further boost performance.
\end{abstract}

\IEEEpeerreviewmaketitle

\section{Introduction}

\IEEEPARstart
{C}{ross-view} geo-localization is to retrieve the most relevant images from different platforms, which could be applied to many fields, such as accurate delivery, autonomous driving, robot navigation, event detection, and so on~\cite{shi_spatial-aware_nodate,zheng_university-1652_nodate,workman_location_2015,liu_lending_2019}. 
For instance, given a drone-view image, 
the system intends to search images of the same location in the candidate images of the satellite. The satellite-view images are automatically annotated with geo-tags. Obtaining the true-match satellite-view image, we could localize the building in the drone-view image. Besides, the image-based cross-view geo-localization can facilitate the positioning devices, \eg, GPS, to provide a more robust and accurate result.
\par
\begin{figure}[htbp]
  \centering
  \includegraphics[width=1\linewidth]{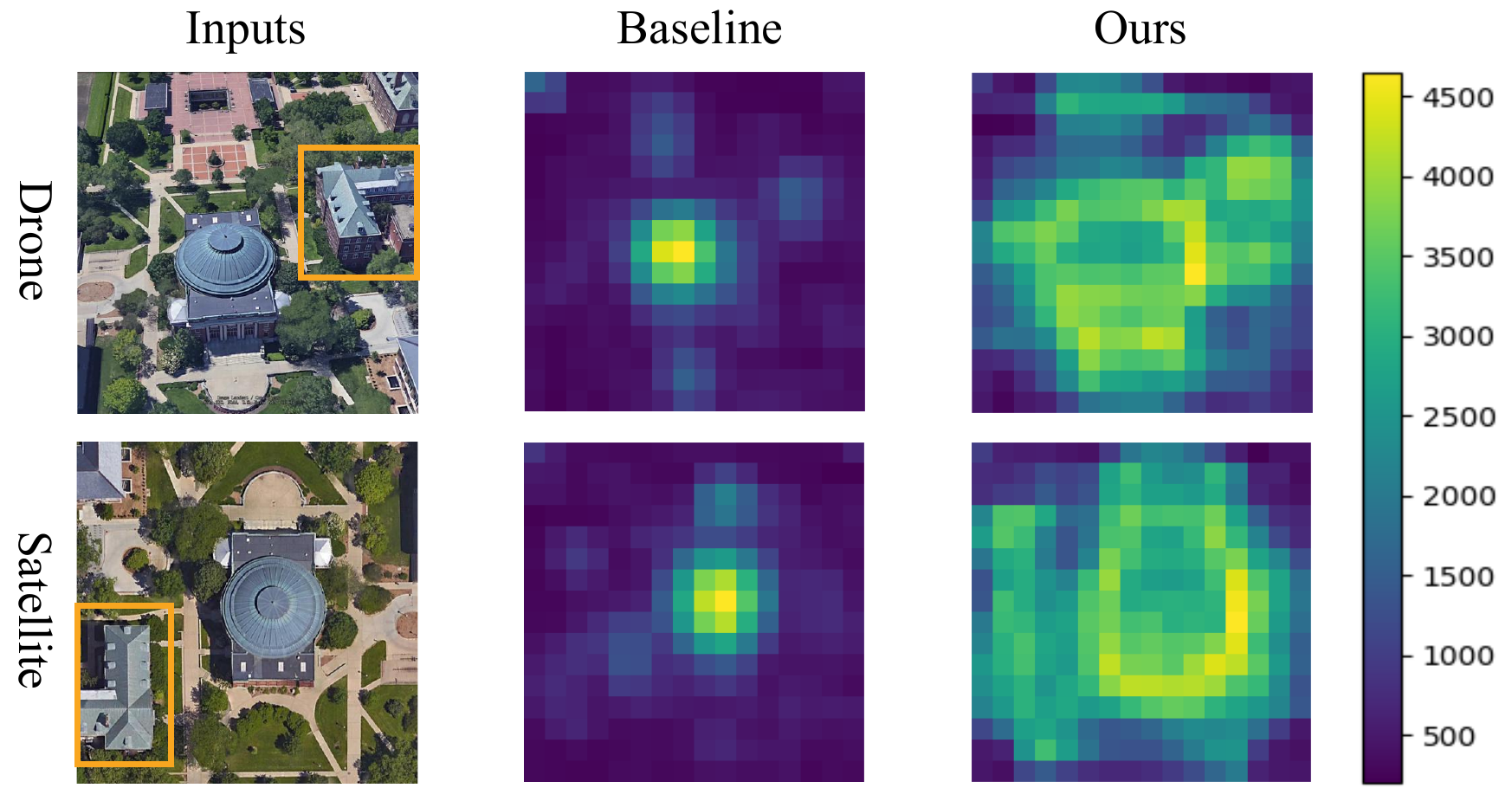}
  \caption{Difference of the activation maps generated by the baseline method~\cite{zheng_university-1652_nodate} and our method. The first column shows two input images from different platforms, \ie, satellite and drone, with the same geo-tag. We observe that the contextual information, such as the neighbor building in the yellow box, can be used as an auxiliary clue to facilitate the cross-view image-based geographic localization. 
  In the second column, we visualize the activation map of the baseline model~\cite{zheng_university-1652_nodate}. We could observe that the baseline method~\cite{zheng_university-1652_nodate} activates only the patterns at the center geographic target, while our method activates more contextual information around the center geographic target (see the third column). ${}^{\dag}$: \textbf{The baseline method is a three-branch network with ResNet-50~\cite{he2016deep} as the backbone, and the model is optimized by the instance loss~\cite{zheng2020dual}.}
  }
  \label{fig:1}
\end{figure}

In recent years, cross-view geo-localization has obtained a significant development due to the advance in deep learning. 
Most works~\cite{shi_spatial-aware_nodate,liu_lending_2019,shi_optimal_nodate,Shi_2020_CVPR,hu_cvm-net_2018} explore the deep neural network with the metric learning to learn the discriminative feature. Specifically, the network is to learn one feature space that brings matched image pairs closer and pushes non-matched pairs far apart~\cite{hadsell2006dimensionality,deng2018image,schroff2015facenet}. The attention mechanism and orientation information are also widely used in the network design~\cite{shi_spatial-aware_nodate,fu2019sta,liu_lending_2019}. However, most existing methods only consider the global information via pooling functions, ignoring the contextual information (see Figure \ref{fig:1}).  

Generally, the aerial-view platform, \eg, drone or satellite, captures the scene image with a wide angle. When the platform acquires a geographic target, the contextual information around the target is also captured as a by-product. 
When existing works usually ignore such information, we argue that the contextual information provides a key clue for cross-view geo-localization. 
For instance, when there is no apparent difference between two geographic targets, such as two straight roads, the human visual system is challenging to identify the true-match target. However, the task is much easier with the help of contextual information, \eg, neighbor houses. Mining and utilizing the contextual information in the image can improve the accuracy of the cross-view geo-localization.  
\par
\begin{figure}[htp]
  \centering
  \includegraphics[width=1\linewidth]{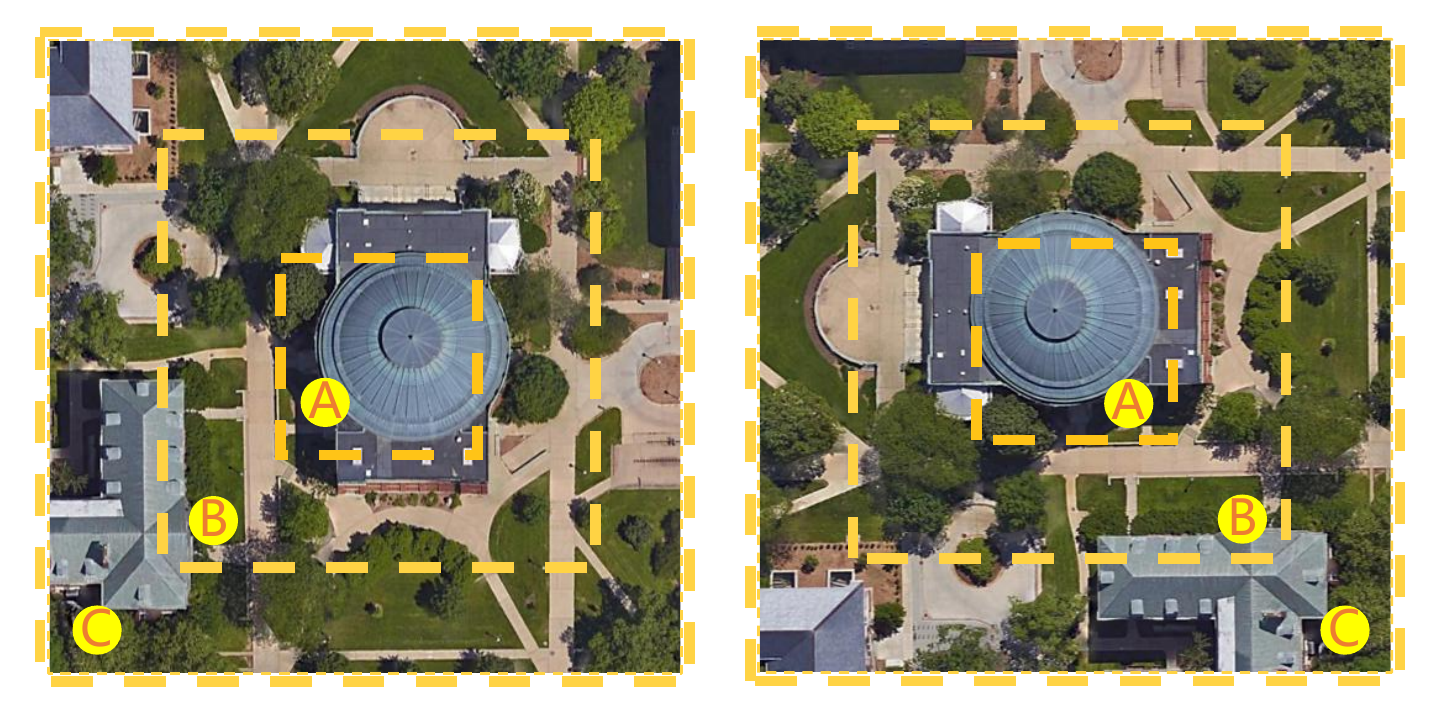}
  \caption{The simplified diagram of our partition strategy, which is invariant to the rotation. The region of part A represents the geographic target in the center. According to the distance from the geographic target, the region of part B can be viewed as the first hierarchical contextual information, and the region of part C is the second hierarchical contextual information. 
  }
  \label{fig:part_strategy}
\end{figure}

Our work is inspired by the procedure that the human visual system interprets and matches the same scene of different viewpoints~\cite{rensink2000dynamic,corbetta2002control,zheng2020vehiclenet}.
When recognizing a geography scene of two different platforms, the human visual system generally adopts a hierarchical processing manner to improve the accuracy of judgement. Specifically, the human visual system first pays attention to whether the same geographic target is contained in different viewpoint scenes. Then, the human visual system will check the contextual information around the geographic target to verify the correctness of the match. When there is no remarkable landmark, people usually resort to the map to find discriminative neighbor areas. Imitating the process mentioned above, we design a Local Pattern Network (LPN), which is an effective way to explicitly explore the contextual information in an end-to-end learning manner. Specifically, we divide the high-level feature into several parts in a square-ring partition, as shown in Figure \ref{fig:part_strategy}. Because the geographic target is generally located in the center of the image with the contextual information surrounded. Our partition method can obtain not only the geographic target information (the region of A) but also several contextual-information parts (the region of B and C) with different distances from the geographic target. Therefore, we can explicitly exploit contextual information to optimize LPN. We also observe that our partition strategy is robust to the image rotation in nature. For instance, when rotating the left image in Figure \ref{fig:part_strategy} as the right image, the three regions (A, B, and C) still contain the same semantic information as corresponding regions of the left image. Therefore, the network designed according to the square-ring manner has good scalability to image rotation.
To verify the effectiveness of the proposed method, we conduct experiments on three public datasets, \ie, University-1652~\cite{zheng_university-1652_nodate}, CVUSA~\cite{zhai_predicting_2017} and CVACT~\cite{liu_lending_2019}. LPN achieves the Recall@1 accuracy of 75.93\% for the task of drone-view target localization (\textbf{Drone $\rightarrow$ Satellite}) and Recall@1 accuracy of $86.45\%$ for the task of drone navigation (\textbf{Satellite $\rightarrow$ Drone}), which is higher than the baseline work~\cite{zheng_university-1652_nodate} by $17.44\%$ and $15.27\%$ respectively. Similar results are also observed on CVUSA and CVACT. Compared with the baseline model~\cite{zheng_university-1652_nodate}, the Recall@1 accuracy increases from $43.91\%$ to $79.69\%$ (+$35.78\%$) on CVUSA and $31.20\%$ to $73.85\%$ (+$42.65\%$) on CVACT.
Besides, the proposed method is complementary to most previous works. The proposed method could be easily fused with the state-of-art methods, \ie, SAFA~\cite{shi_spatial-aware_nodate}, and boost the performance from $89.84\%$ Recall@1 accuracy to $92.83\%$ (+$2.99\%$) Recall@1 accuracy on CVUSA and $81.03\%$ Recall@1 accuracy to $83.66\%$ (+$2.63\%$) Recall@1 accuracy on CVACT.
\par
In summary, the main contributions of this paper are as follows:
\begin{itemize}
\setlength\itemsep{0em}
\item We propose a simple and effective model, called Local Pattern Network (\textbf{LPN}). Different from existing works, LPN explicitly takes contextual patterns into consideration and leverages the surrounding environment around the target building. Specifically, the model deploys the square-ring partition strategy and learns contextual information in an end-to-end manner. 
\item We demonstrate the effectiveness of our method on three prevailing cross-view geo-localization datasets, \ie, University-1652~\cite{zheng_university-1652_nodate}, CVUSA~\cite{zhai_predicting_2017} and CVACT~\cite{liu_lending_2019}. Our method outperforms the strong baseline on both benchmarks by a large margin. Furthermore, we show that the proposed method is complementary to existing works, and can be fused with the state-the-art approaches to further boost the  performance.
\end{itemize}
\par
We organize the rest of this paper as follows. In Section \ref{related_work}, we briefly introduce some of the relevant works. Section \ref{proposed method} presents our designed LPN in detail. Experimental results are presented in Section \ref{experiment} and followed by the conclusion in Section \ref{conclusion}.

\section{Related Work}\label{related_work}
In this section, we briefly review related previous works, including deep cross-view geo-localization and part-based representation learning.

\subsection{Deep Cross-view Geo-localization}
Cross-view geo-localization has been attracting more attention in recent years due to a large number of potential applications. Some pioneering approaches~\cite{SemanticCM,lin2013cross,senlet2011framework,bansal2011geo} focus on extracting hand-crafted features. Inspired by the great success of the deep convolutional neural networks (CNNs) on ImageNet, researchers resort to the deeply-learned feature in recent years. 
Workman \etal~\cite{workman_location_2015} are among the first attempts to utilize a pre-trained CNN to extract features for the cross-view localization task. They demonstrate that features from the high-level layer of CNN contain semantic information about the geographic location. To take one step further, Workman \etal ~\cite{workman_wide-area_2015} fine-tune the pre-trained network by reducing the feature distance between pairs of ground-level images and aerial images, yielding better performance. 
Inspired by the face verification approaches, Lin \etal~\cite{lin_learning_2015} adopt a modified Siamese Network~\cite{chopra2005learning}, which optimizes network parameters by the contrastive loss~\cite{hadsell2006dimensionality,deng2018image}. 
Zhai \etal~\cite{hu_cvm-net_2018} plug the NetVLAD~\cite{arandjelovic2016netvlad} into a Siamese-like architecture, making image descriptors robust against large viewpoint changes. 
Liu \etal~\cite{liu_lending_2019} stress the importance of orientation information and encode corresponding coordinate information into the network to boost the discrimination of the feature. 
In a recent work, Shi \etal~\cite{shi_optimal_nodate} use the spatial layout information to make up the shortcoming of the global aggregation step in feature extraction. Furthermore, Shi \etal~\cite{shi_spatial-aware_nodate} improve the performance of cross-view geo-localization through domain alignment and spatial attention mechanism. Besides, DSM~\cite{Shi_2020_CVPR} considers a limited Field of View setting and adopts a dynamic similarity matching module to align the orientation of cross-view images. 
Another line of works considers the metric learning and designs different training objectives to learn the discriminative representation.
Vo \etal~\cite{vo_localizing_2017} design an orientation regression loss, yielding the performance improvement. 
Hu~\etal~\cite{hu_cvm-net_2018} employ a weighted soft margin ranking loss, which not only speeds up the training convergence but also improves the retrieval accuracy. Different from adopting metric learning loss (\ie, contrastive loss~\cite{hadsell2006dimensionality,deng2018image} and triplet loss~\cite{schroff2015facenet,li2020hierarchical}), Zheng~\etal~\cite{zheng_university-1652_nodate} regard the cross-view image retrieval as a classification task. They apply the instance loss~\cite{zheng2017unlabeled,zheng2020dual} to optimize the network and has achieved a competitive result. 
However, these methods usually concentrate on exploring the global information but ignore the contextual information as shown in Figure~\ref{fig:1}. Different from existing works, the proposed method intends to take advantage of the neighbor areas. We deploy the feature-level partition strategy, which facilitates the end-to-end learning on the contextual information.   

\subsection{Part-based Representation Learning}
The local feature has been widely studied in the design of hand-crafted algorithms~\cite{amit2007pop,crandall2005spatial,fergus2003object,leibe2008robust,weber2000towards}. Ojala~\etal~\cite{LBP} propose a local binary pattern (LBP) descriptor to extract the rotation-invariant feature. Lowe~\etal~\cite{SIFT} develop a Scale Invariant Feature Transform (SIFT) descriptor for the image-based match. SIFT is invariant to translations, rotation, and scaling transformations by summarizing description of the local image structures in a local neighborhood around each interest point. 
In the spirit of the conventional part-based descriptor, some researchers also explore the local pattern learning in the deep-leaned models.
One line of works divides the features based on an extra estimator, such as landmark detection, human pose estimated, and human parsing. 
Spindle Net~\cite{zhao_spindle_2017} leverages the landmark points of the human body to obtain semantic features from different body regions. 
Xu \etal~\cite{xu_attention-aware_2018} propose a pose-guide part attention module to learn a confidence map. 
Guo \etal~\cite{guo_beyond_2019} acquire the accurate human part-aligned representation by the human parsing model to enhance the robustness of the feature.
Another line of works does not need an extra pose estimator and deploys a coarse part alignment, such as horizontal matching. Li \etal~\cite{li_learning_2017} capture the three parts information corresponding to the head-shoulder, upper body, and lower body by Spatial Transformer Network (STN)~\cite{zheng2018pedestrian,jaderberg_spatial_2015}. Zhao \etal~\cite{zhao_deeply-learned_2017} utilize the attention mechanism to learn aligned part information from the input image automatically. A strong Part-based Convolutional Baseline (PCB)~\cite{sun_beyond_2018} shows a uniform partition strategy to extract high-level features. Then, by correcting within-part inconsistency of all the column vectors according to their similarities to each part, the performance of this work becomes better. Currently, some state-of-the-art works~\cite{sun_dissecting_2019,zhong_invariance_2019,song_generalizable_2019,fu2019horizontal} extend the PCB with more partitions or optimization losses. 
Our work also studies a part-based representation learning on the convolutional layer, but is different in two aspects: 
Different from works of the first line~\cite{xu_attention-aware_2018,zhao_spindle_2017,zheng_pose-invariant_2019,su_pose-driven_2017,wei_glad_2019}, the proposed method does not need an extra part estimator. Different from works of the second line~\cite{zheng2018pedestrian,sun_beyond_2018,sun_dissecting_2019,zhong_invariance_2019}, our partition method makes the network have good scalability to image rotation (see Figure \ref{fig:part_strategy}).

\begin{figure*}[htb]
\begin{center}
    \includegraphics[width=1\linewidth]{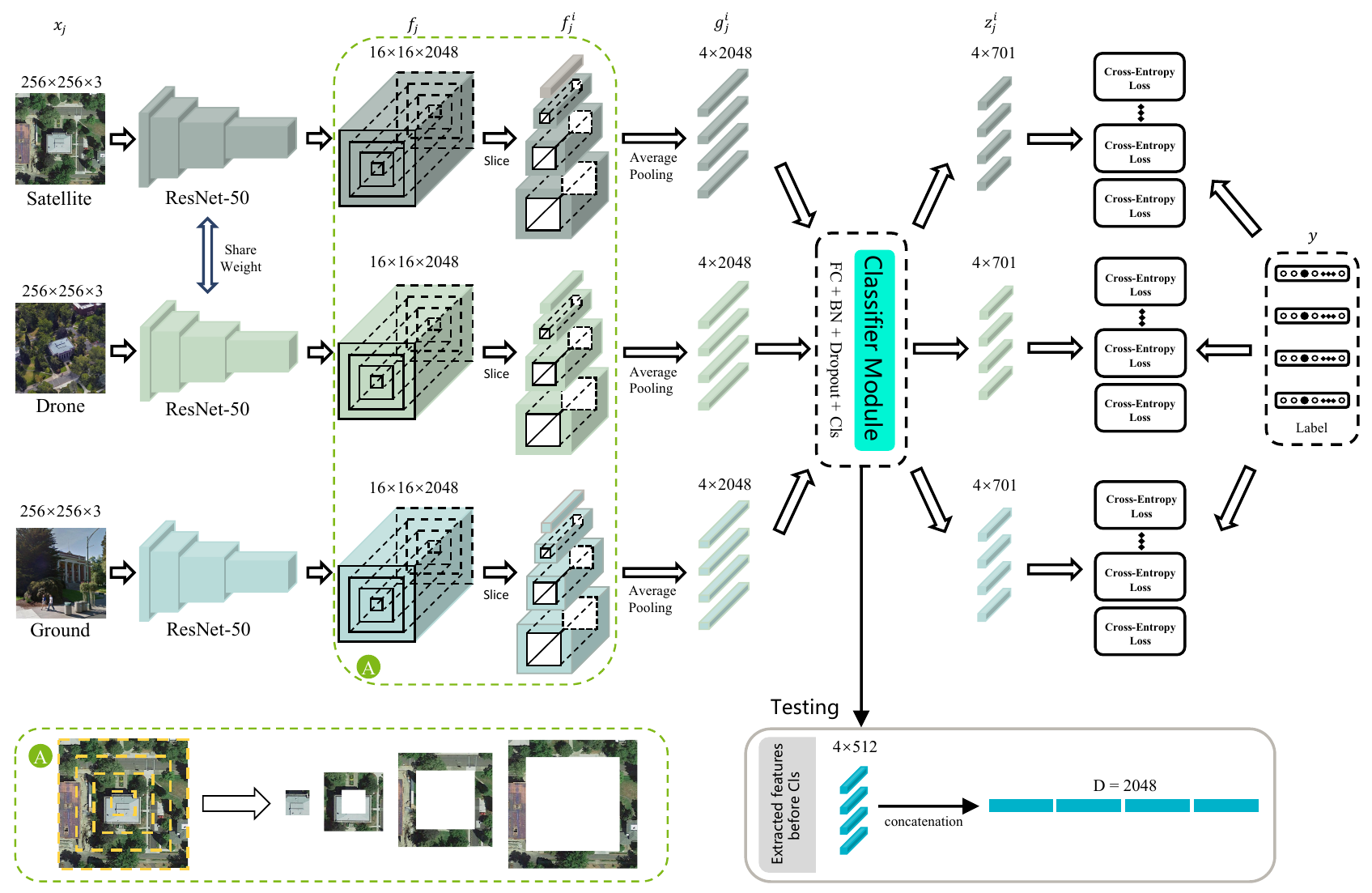}
\end{center}
\caption{Overview of the proposed LPN framework. 
Given one input image, we first extract feature maps. Since we study the cross-view geo-localization, the input image can be from different platforms. The proposed LPN contains three branches, \ie, the satellite-view branch, the drone-view branch and the ground-view branch, respectively, to deal with different kinds of inputs. 
The satellite-view branch and the drone-view branch share weights since images from the satellite view and the drone view have similar patterns. 
Then, the output feature maps from each branch are sliced according to the square-ring partition strategy. Next, the average pooling layer is used to transform each part-level feature maps into a column feature descriptor. Finally, all these feature descriptors are fed into a classifier module to get prediction vectors. In addition to the classification layer (Cls), the classifier module also contains other three type layers, which are the fully-connected layer (FC), the batch normalization layer (BN), 
and the dropout layer (Dropout). 
During training, we leverage the classifier module to predict the geo-tag of each part. 
The network is optimized by minimizing the sum of the cross-entropy losses over all parts.
When testing, we obtain the part-level image representation before the classification layer in the classifier module. Then we concatenate part-level features as the final visual descriptor of the input image, and the dimension of the feature is $2048$.  
In (A) (a green dotted line), we show the square-ring partition strategy. Note that here we display the framework for the University-1652 dataset of input data from three platforms. For two-view datasets, \eg, CVUSA, we use two CNN branches. 
}\label{fig:main}
\end{figure*}

\section{Proposed Method}\label{proposed method}
In this section, we introduce the Local Pattern Network (LPN) (see Figure \ref{fig:main}). We first illustrate the network architecture for feature extraction, followed by the partition strategy for feature maps and the optimization objective. Finally, we provide a discussion on our intuition and special cases for different datasets. 

\textbf{Problem formulation.} Given one geo-localization dataset, we denote the input image as $x$, and $y$ represents the corresponding label. We apply the subscript $j$ to denote the platform where the data $x_j$ is collected, and $j \in \{1, 2, 3\}$. 
In particular, $x_1$ denotes the sample from  the satellite view, $x_2$ denotes the drone-view data, and $x_3$ denotes the ground-view image. The label
$y \in [1, C]$, where $C$ indicates the number of categories. For instance, a dataset includes 701 buildings and each building contains multiple images. We number 701 buildings into 701 different indexes. Each index represents a category, \ie, the label $y \in [1, 701]$. For cross-view geo-localization, we intend to learn one mapping function, which could project images from different platforms to one shared semantic space. The images of the same location are close, while the images from different location are apart from each other. 

\begin{figure}[htbp]
  \centering
  \includegraphics[width=1\linewidth]{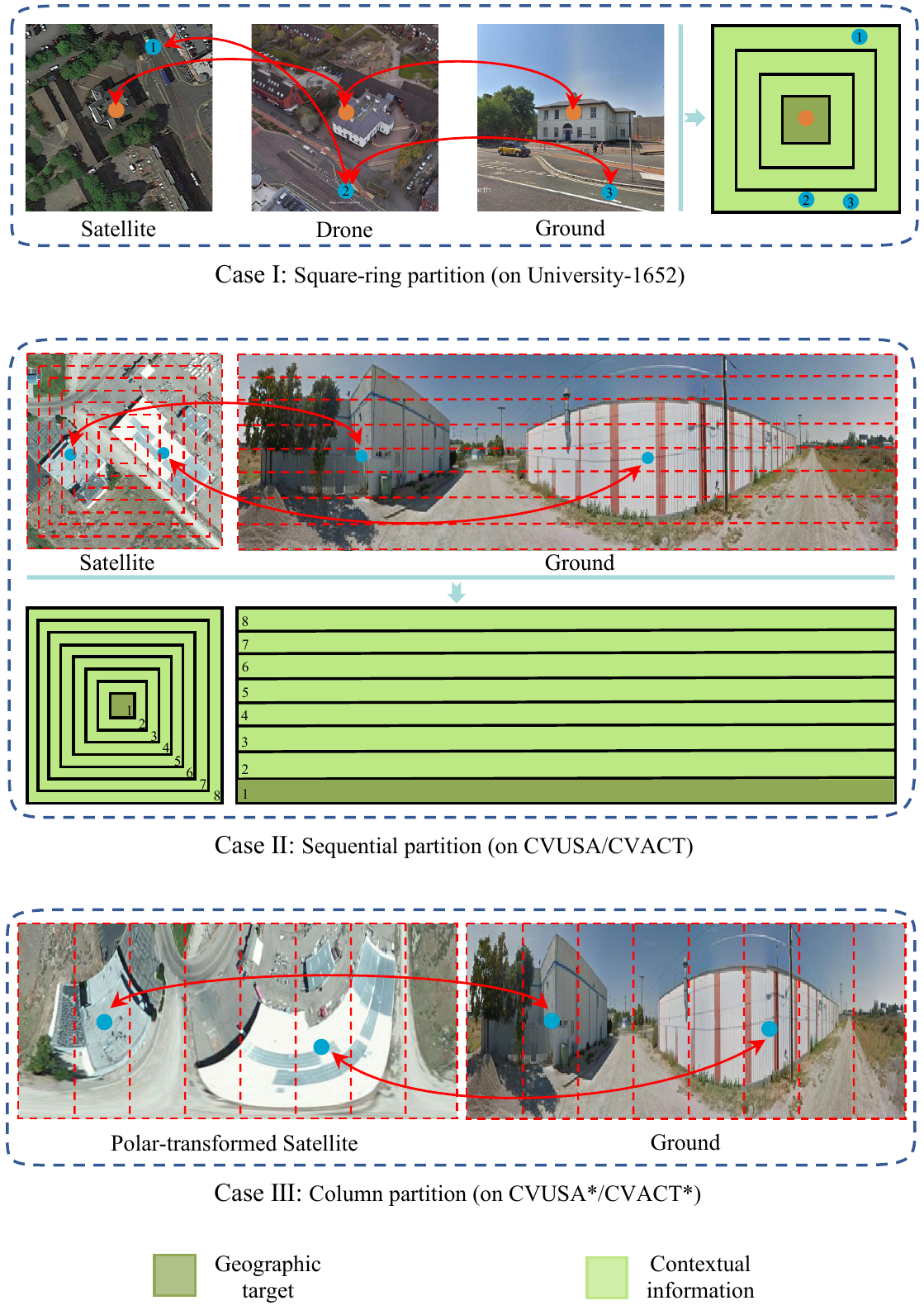}
  \caption{Illustration of the square-ring partition (Case \uppercase\expandafter{\romannumeral1}), the sequential partition (Case \uppercase\expandafter{\romannumeral2}) and the column partition (Case \uppercase\expandafter{\romannumeral3}). \textbf{The sequential partition strategy and the column partition strategy are two special cases of the square-ring partition strategy.} The sequential partition considers the geometric correspondence for matching  satellite-and-ground panorama image pair~\cite{shi_spatial-aware_nodate,Shi_2020_CVPR}. The column partition directly splits the feature maps vertically. All three partition strategies exploit the contextual information and achieve the spatial alignment of each part. The square-ring partition strategy is suitable for processing images that the contextual information is distributed around the geographic target, such as University-1652. When localizing panoramic image and the orientation of different view images is aligned, the sequential partition has a higher priority, such as CVUSA and CVACT. The column partition enables a fine-grained spatial alignment for pre-processed image pairs on CVUSA$^*$ and CVACT$^*$~\cite{shi_spatial-aware_nodate,shi_optimal_nodate} whose orientation and semantic information distribution are roughly aligned.
 }
  \label{fig:visual}
\end{figure}

\subsection{Local Pattern Network}
\textbf{Feature extraction.}
The proposed model, \ie, Local Pattern Network (LPN), contains three branches, which extends from the Siamese network~\cite{chopra2005learning}. From top to bottom in Figure~\ref{fig:main}, the three branches are the satellite-view branch, the drone-view branch and the ground-view branch respectively. LPN can deploy various network architectures as backbones to extract features, such as VGG~\cite{vgg}, and ResNet~\cite{he2016deep}. For illustration, we choose ResNet-50~\cite{he2016deep} as the network architecture of each branch if not specified. ResNet-50 contains five blocks named conv1, conv2, conv3, conv4, conv5, one average pooling layer, and one fully connected layer. 
We remove the final average pooling layer and the fully connected layer, and obtain intermediate feature maps for subsequent partition processing. 
Following ~\cite{zheng_university-1652_nodate}, we share weights between the satellite-view branch and the drone-view branch, since input images of both branches are from the aerial viewpoint. Three branches have the same feature extraction manner. Specifically, given an input image of size $256 \times 256$, we can acquire feature maps with the shape of $16 \times 16 \times 2048$ in each branch. We denote this function as $\mathcal{F}_{backbone}$, and the process of feature extraction can be formulated as:
\begin{align}\label{extract}
    f_{j} &= \mathcal{F}_{backbone}(x_j),
\end{align}
where $f_{j}$ stands for the extracted feature map of the input image $x_j$. 

\textbf{Feature partition strategy.}\label{partition}
To explicitly take advantage of contextual information, we apply the square-ring partition strategy to divide feature maps. 
We observe that the geographic target is generally distributed in the center of the image, and the contextual information is radiantly distributed around. 
Based on this assumption of semantic information distribution, the center of the square-ring partition can be approximately aligned at the center of the feature maps. As shown in Figure~\ref{fig:main} (A) (green box), we separate images into four parts according to the distance to the image center. In practice, we separate the global feature maps $f_j$ to four feature parts $f_j^i (i\in\{1,2,3,4\})$. The superscript $i$ represents the $i$-th part from the center. 
Then we apply the average pooling layer to transform each part $f_j^i$ with different shapes into a 2048-dim part feature $g_j^i$. The process can be formulated as:
\begin{align}\label{f_ji}
    f_{j}^i &= \mathcal{F}_{slice}(f_{j}, i),
\end{align}
\begin{align}\label{g_ji}
    g_{j}^i &= \mathcal{A}{vgpool}(f_{j}^i),
\end{align}
where $\mathcal{F}_{slice}$ indicates the square-ring partition, and $\mathcal{A}{vgpool}$ represents the average pooling operation.

\textbf{Optimization objective.}
Now we have obtained part features from different sources. Since the features are extracted from different branches, they may have different distribution, which could not be directly used for matching. 
To solve this limitation, we set up a mapping function that maps features of all sources into one shared feature space. 
In this shared space, features of the same geo-tag will have a closer distance, while features of different geo-tags are apart from each others.
This classifier module consists of following layers: a fully connected layer (FC), 
a batch normalization layer (BN), 
 a dropout layer (Dropout), 
and a classification layer (Cls), which is a fully-connected layer. 
The classifier module is deployed to predict the geo-tag of each image based on part features. 
Given the part features $g_j^i$ as the input, the classifier module outputs a column vector $z_j^i$. The dimension of $z_j^i$ equals the number of geo-tag categories $C$.
\begin{align}\label{z}
    z_{j}^i &= \mathcal{F}_{classifier}(g_j^i).
\end{align}
The cross-entropy loss could be formulated as:
\begin{align}\label{softmax}
    \hat{p}(y|x_j^i) &= \frac{exp(z_j^i(y))}{\sum_{c=1}^{C}exp(z_j^i(c))},
\end{align}
\begin{align}\label{opt}
    Loss &= \sum_{i,j}-log(\hat{p}(y|x_j^i))),
\end{align}
where $z_j^i(y)$ is  the logit score  of the ground-truth geo-tag  $y$. We apply the softmax function (Equation~\ref{softmax}) to obtain the normalized probability score $\hat{p}(y|x_j^i)$ in $[0,1]$. $\hat{p}(y|x_j^i)$ is the predicted probability that $x_j^i$ belongs to the geo-tag  $y$. 
In Equation \ref{opt}, we accumulate the losses on the image of different parts and different platforms to optimize the whole network. 

\subsection{Discussion}\label{discussion}
Our method is inspired by the mechanism of the human visual system on matching images of different viewpoints. In the ancient time, people compare the map and the surrounding environment to know where they are. The contextual information plays an important role. 
Nowadays, cameras on different platforms typically use wide-angle lenses to obtain complete geographic targets, and the contextual information around the geographic target is also collected in the image. We argue that the contextual information, as a by-product, can facilitate the discriminative representation learning. For example, the neighbor building also could help to predict the target location.
Instead of dividing images in the pixel level, we split the feature maps in practice, which could not only improve the efficiency but also enable the larger receptive fields as well as the part alignment. The square-ring partition strategy is also robust to the rotation variants. Case \uppercase\expandafter{\romannumeral1} (see Figure~\ref{fig:visual}) shows the application of the square-ring partition strategy on three images of different views in University-1652, \ie, satellite view, drone view and ground view. The orientation of these three-view images is not aligned. However, we can observe that the geographic target (orange point) is generally located in the image center. Because of the random orientation of three-view images, the contextual information with the same semantics (blue point) may not be distributed in the same orientation but can be located in the same part.
We note that the sequential partition strategy is a special case of the square-ring partition strategy. The sequential partition strategy takes into account the geometric correspondence~\cite{shi_spatial-aware_nodate,Shi_2020_CVPR} for a north aligned satellite-and-ground panorama image pair, such as data on CVUSA~\cite{zhai_predicting_2017}/CVACT~\cite{liu_lending_2019}. \textbf{Specifically, we apply the square-ring partition to satellite images and the row partition to ground panoramas.}
As shown in Case \uppercase\expandafter{\romannumeral2} of Figure~\ref{fig:visual}, image pairs on CVUSA/CVACT have different visual appearances. But the same semantic information (\eg, the blue point pair) from the true-matched image pair can still be roughly located in the same part of the divided feature maps. Besides, the column partition is also a variation of our method. The column partition can be adapted when the orientation and the spatial semantics of the matching image pair are roughly aligned. For example, images have been pre-processed by Optimal Transport theory~\cite{shi_optimal_nodate} or the polar transform~\cite{shi_spatial-aware_nodate,Shi_2020_CVPR}. Case \uppercase\expandafter{\romannumeral3} (see Figure~\ref{fig:visual}) provides an example that the column partition is applied to a polar-transformed satellite image and a ground panorama.

\setlength{\tabcolsep}{5pt}
\begin{table}[htb]
\small
\caption{
Statistics of three different test sets, including the image number of query set and gallery set for different geo-localization tasks.
}
\begin{center}
\begin{tabular}{l|cc|cc}
\hline
\multicolumn{1}{c|}{\multirow{3}{*}{Dataset}} & \multicolumn{4}{c}{Task} \\
\cline{2-5}
& \multicolumn{2}{c|}{Drone $\rightarrow$ Satellite} & \multicolumn{2}{c}{Satellite $\rightarrow$ Drone}\\
& Query & Gallery & Query & Gallery \\
\shline
University-1652~\cite{zheng_university-1652_nodate} & 37,855 & 951 & 701 & 51,355 \\
\hline
\cline{2-5}
& \multicolumn{2}{c|}{Ground $\rightarrow$ Satellite} & \multicolumn{2}{c}{Satellite $\rightarrow$ Ground}\\
& Query & Gallery & Query & Gallery \\
\shline
CVUSA~\cite{zhai_predicting_2017} & 8884 & 8884 & 8884 & 8884 \\
CVACT\_val~\cite{liu_lending_2019} & 8884 & 8884 & 8884 & 8884 \\
\hline
\end{tabular}
\end{center}
\label{table:statistic}
\end{table}

\section{Experiment}\label{experiment}
We first introduce three large-scale cross-view geo-localization datasets, two small-scale landmark retrieval datasets and the evaluation protocol. Then Section \ref{implementation detail} describes the implementation detail. We provide the comparison with the state of the arts in Section \ref{sota}, followed by the ablation study in Section \ref{ablation}.
\subsection{Datasets and Evaluation Protocol}
We mainly train and evaluate our method on three large-scale geo-localization datasets, \ie, University-1652~\cite{zheng_university-1652_nodate}, CVUSA~\cite{zhai_predicting_2017} and CVACT~\cite{liu_lending_2019}. Table \ref{table:statistic} shows the image number of query and gallery sets for testing different tasks using these three datasets.
\par
\textbf{University-1652}~\cite{zheng_university-1652_nodate} is a multi-view multi-source dataset containing satellite-view data, drone-view data and ground-view data. 
It collects 1652 buildings of 72 universities around the world. 
The training set includes 701 buildings of 33 universities, and the testing set includes the other 951 buildings of the rest 39 universities. \textbf{There are no overlapping universities in the training and test set.} 
Since some buildings do not have enough ground-view images to cover different aspects of these buildings, the dataset also provides an additional training set. Images in the additional training set are collected from the Google Image, and they have a similar view as the ground-view images. Therefore, the additional training set can be used as a supplement of the ground-view images. 
The dataset is employed to study two new tasks, \ie{}, drone-view target localization (Drone $\rightarrow$ Satellite) and drone navigation (Satellite $\rightarrow$ Drone). There are 701 buildings with 50,218 images for training. 
In the drone-view target localization task (Drone $\rightarrow$ Satellite), there are 37,855 drone-view images in the query set and 701 true-matched satellite-view images and 250 satellite-view distractors in the gallery. There is only one true-matched satellite-view image under this setting. In the drone navigation task (Satellite $\rightarrow$ Drone), there are 701 satellite-view query images, and 37,855 true-matched drone-view images and 13,500 drone-view distractors in the gallery. There are multiple true-matched drone-view images under this setting.
\par
\textbf{CVUSA}~\cite{zhai_predicting_2017} provides the data collected from two views, \ie, the ground view and the satellite view. Specifically, it contains 35,532 ground-and-satellite image pairs for training and 8884 image pairs for testing. All ground-view panoramic images are collected from Google Street View. Meanwhile, corresponding satellite-view images are downloaded from Microsoft Bing Maps. 
\par
\textbf{CVACT}~\cite{liu_lending_2019} is a large-scale cross-view dataset. Same as CVUSA, CVACT provides 35,532 ground-and-satellite image pairs for training, and ground-view images are panoramas. Besides, CVACT provides a validation set with 8884 image pairs named CVACT\_val and a testing set with 92,802 image pairs denoted as CVACT\_test. A query image only has one true-matched image in the gallery for CVACT\_val, while for CVACT\_test, a query image may correspond to several true-matched images in the gallery.
\par
\textbf{Oxford5k~\cite{philbin2007object} \& Paris6k~\cite{philbin2008lost}} are two prevailing landmark retrieval datasets collected from Flickr. Oxford5k consists of 5062 images that belong to 11 different Oxford landmarks, and Paris6k contains 6412 images of 12 particular Paris buildings. There are 55 query images in Oxford5k and 12 queries in Paris6k. 
\par
\textbf{Evaluation protocol.} 
In our experiments, we use the Recall@K (\textbf{R@K}) and the average precision (\textbf{AP}) to evaluate the performance of our model. R@K represents the proportion of correctly matched images in the top-K of the ranking list. A higher recall score shows a better performance of the network. We also calculate the
area under the Precision-Recall curve, which is known
as the average precision (AP), which reflects the
precision and recall rate of the retrieval performance. 
\subsection{Implementation Details} \label{implementation detail}
We employ the ResNet-50~\cite{he2016deep} with pre-trained weights on ImageNet~\cite{5206848} to extract visual features. Following ~\cite{zheng_university-1652_nodate}, we modify the stride of the second convolutional layer and the last down-sample layer in conv5\_1 of the ResNet-50 from 2 to 1. The newly-added layers in LPN, \ie, the classifier module, are initialized with \textit{kaiming initialization}~\cite{kaiming_init}. We resize each input image to a fixed size of 256 $\times$ 256 pixels during training and testing. In training, we employ random cropping and flipping to augment the input data. For the optimizer, we adopt stochastic gradient descent (SGD) with momentum 0.9 and weight decay 0.0005 with a mini-batch of 32. The initial learning rate is 0.001 for backbone layers and 0.01 for the new layers. We train our model for 120 epochs, and the learning rate is decayed by 0.1 after 80 epochs. 
During testing, we utilize the Euclidean distance to measure the similarity between the query image and candidate images in the gallery. Our model is implemented on Pytorch~\cite{paszke2019pytorch}, and all experiments are conducted on one NVIDIA RTX 2080Ti GPU.

\subsection{Comparison with the State-of-the-arts}\label{sota}
\textbf{Results on University-1652.} 
As shown in Table \ref{table:university1652}, we compare the proposed method with other competitive approaches on University-1652. 
The proposed LPN has achieved $74.18\%$ Recall@1 accuracy and $77.39\%$ AP on Drone $\rightarrow$ Satellite and $85.16\%$ Recall@1 accuracy and $73.68\%$ AP on Satellite $\rightarrow$ Drone without using the additional Google training data. 
The performance has surpassed the reported result of other competitive methods such as~\cite{lin_learning_2015,workman_wide-area_2015,chechik2009large,deng2018triplet,hu_cvm-net_2018,liu_lending_2019}, and the proposed method outperforms the best method, \ie, instance loss~\cite{zheng_university-1652_nodate} by a large margin, \ie, about $14\%$ AP improvement.
If the extra training data, \ie, noisy data collected from Google Image, is added into the training set \cite{zheng_university-1652_nodate}, we could further boost the retrieval performance. In the drone-view target localization task (Drone $\rightarrow$ Satellite), the accuracy of Recall@1 increases from $74.18\%$ to $75.93\%$ and the value of AP goes up from $77.39\%$ to $79.14\%$; in the drone navigation task (Satellite $\rightarrow$ Drone), the accuracy of Recall@1 increases from $85.16\%$ to $86.45\%$ and the value of AP raises from $73.68\%$ to $74.79\%$. For the drone-view target localization task, there are 951 satellite-view images in the gallery. To make this retrieval task more challenging, we add 8884 satellite-view images collected from the testing set of CVUSA into the gallery of University-1652 as the distractors. Although the distractors would decrease the overall performance, indicated by Rank@1 and AP accuracy, the results are still competitive. This demonstrates the robustness of our proposed method against distractors.

\par
\setlength{\tabcolsep}{0.8pt}
\begin{table}[htb]
\small
\caption{
Comparison with the state-of-the-art results reported on University-1652. $M$ stands for the margin of the triplet loss. (w/o Google) indicates that the extra training set collected from Google Image is not deployed in training phase. (w/ CVUSA distractors) denotes that all satellite-view images collected from the testing set of CVUSA are added into the satellite-view gallery of University-1652 as the distractors.} 
\begin{center}
\begin{tabular}{l|cc|cc}
\hline
\multicolumn{1}{c|}{\multirow{3}{*}{Method}} & \multicolumn{4}{c}{University-1652} \\
\cline{2-5}
& \multicolumn{2}{c|}{Drone $\rightarrow$ Satellite} & \multicolumn{2}{c}{Satellite $\rightarrow$ Drone}\\
& R@1 & AP & R@1 & AP \\
\shline
Instance Loss~\cite{zheng_university-1652_nodate} & 58.49 & 63.31 & 71.18 & 58.74 \\
Contrastive Loss~\cite{lin_learning_2015} & 52.39 & 57.44 & 63.91 & 52.24 \\
Triplet Loss ($M=0.3$)~\cite{chechik2009large} & 55.18 & 59.97 & 63.62 & 53.85 \\
Triplet Loss ($M=0.5$)~\cite{chechik2009large} & 53.58 & 58.60 & 64.48 & 53.15 \\
Soft Margin Triplet Loss~\cite{hu_cvm-net_2018} & 53.21 & 58.03 & 65.62 & 54.47 \\
\hline
Ours (w/o Google) & 74.18 & 77.39 & 85.16 & 73.68 \\
Ours & \textbf{75.93} & \textbf{79.14} & \textbf{86.45} & \textbf{74.79} \\
Ours (w/ CVUSA distractors) & 70.61 & 73.53 & - & - \\
\hline
\end{tabular}
\end{center}
\label{table:university1652}
\end{table}

\begin{table*}[htp]
\setlength{\tabcolsep}{7pt}
\centering
\small
\caption{
Results on CVUSA, list shows comparisons of various methods. There are two schemes to optimize the network, \ie, instance loss and deep metric learning. Zheng~\etal~\cite{zheng_university-1652_nodate} get the best result in the scheme of instance loss, while, in the deep metric learning scheme, SAFA~\cite{shi_spatial-aware_nodate} is a state-of-the-art work. we observe that through combining our method to these two methods, the off-the-shelf network can achieve a significant performance boost. $^\dag$: The method utilizes extra orientation information as input.
}
\begin{tabular}{l|c|c|cccc|cccc}
\hline
\multicolumn{1}{c|}{\multirow{2}{*}{Method}} & \multirow{2}{*}{Publication} & \multirow{2}{*}{Backbone} & \multicolumn{4}{c|}{CVUSA} & \multicolumn{4}{c}{CVACT\_val}\\ 
\cline{4-11}
                                        & & & R@1 & R@5 & R@10 & R@Top1\% & R@1 & R@5 & R@10 & R@Top1\% \\
\shline
MCVPlaces~\cite{workman_wide-area_2015} & ICCV'15 & AlexNet & - & - & - & 34.40 & - & - & - & - \\
Zhai~\cite{zhai_predicting_2017} & CVPR'17 & VGG16 & - & - & - & 43.20 & - & - & - & -\\
Vo~\cite{vo_localizing_2017}    & ECCV'16 & AlexNet & - & - & - & 63.70 & - & - & - & -\\
CVM-Net~\cite{hu_cvm-net_2018} & CVPR'18 & VGG16 & 18.80 & 44.42 & 57.47 & 91.54 & 20.15 & 45.00 & 56.87 & 87.57\\
Orientation$^\dag$~\cite{liu_lending_2019}   & CVPR'19 & VGG16 & 27.15 & 54.66 & 67.54 & 93.91 & 46.96 & 68.28 & 75.48 & 92.04\\
Zheng~\cite{zheng_university-1652_nodate} & MM'20 & VGG16 & 43.91 & 66.38 & 74.58 & 91.78 & 31.20 & 53.64 & 63.00 & 85.27\\
Regmi~\cite{Regmi_2019_ICCV} & ICCV'19 & X-Fork & 48.75 & - & 81.27 & 95.98 & - & - & - & -\\
Siam-FCANet~\cite{Siam-FCANet} & ICCV'19 & ResNet-34 & - & - & - & 98.30 & - & - & - & -\\
CVFT~\cite{shi_optimal_nodate} & AAAI'20 & VGG16 & 61.43 & 84.69 & 90.94 & 99.02 & 61.05 & 81.33 & 86.52 & 95.93\\
SAFA~\cite{shi_spatial-aware_nodate} & NIPS'19 & VGG16 & 89.84 & 96.93 & 98.14 & 99.64 & 81.03 & 92.80 & 94.84 & 98.17\\
\hline
Ours & - & VGG16 & 79.69 & 91.70 & 94.55 & 98.50 & 73.85 & 87.54 & 90.66 & 95.87\\
Ours & - & ResNet-50 & \textbf{85.79} & \textbf{95.38} & \textbf{96.98} & \textbf{99.41} & \textbf{79.99} & \textbf{90.63} & \textbf{92.56} & \textbf{97.03}\\
CVFT~\cite{shi_optimal_nodate} + Ours & - & VGG16 & 68.20 & 88.00 & 92.69 & 99.30 & 62.90 & 84.14 & 89.11 & 97.22\\
SAFA~\cite{shi_spatial-aware_nodate} + Ours & - & VGG16 & \textbf{92.83} & \textbf{98.00} & \textbf{98.85} & \textbf{99.78} & \textbf{83.66} & \textbf{94.14} & \textbf{95.92} & \textbf{98.41}\\
\hline
\end{tabular}
\label{table:CVUSA}
\end{table*}
\textbf{Results on CVUSA.}
The comparison with other competitive methods on CVUSA is detailed in Table \ref{table:CVUSA}. 
Ground-view images on CVUSA are panoramas, in which, the contextual information is generally distributed on both sides of the geographic target. Basing on the discussion in \ref{discussion}, we deploy the sequential partition strategy to explicitly mine the contextual information on CVUSA (see Figure \ref{fig:visual}). The sequential partition strategy is a specific case of the square-ring partition strategy.
As shown in Table \ref{table:CVUSA}, we could observe two points. 
First, we deploy category recognition as the pretext task to conduct geo-localization on CVUSA. In particular, we regard 35,532 pairs as 35,532 location categories to train the model. The proposed method, whether using VGG16~\cite{vgg} or ResNet-50~\cite{he2016deep} as the backbone, surpasses most existing methods. Specifically, when using VGG16 as the backbone, our method achieves $79.69\%$ R@1, $91.70\%$ R@5, $94.55\%$ R@10 and $98.50\%$ R@Top1\% on CVUSA. Since the feature expression capability of ResNet-50 is powerful than VGG16, our method with ResNet-50 backbone obtains $85.79\%$ R@1, $95.38\%$ R@5, $96.98\%$ R@10 and $99.41\%$ R@Top1\% on CVUSA.
Second, the proposed method is complementary to existing methods. 
For instance, our method can combine with the CVFT~\cite{shi_optimal_nodate} and the SAFA~\cite{shi_spatial-aware_nodate} orthogonally. We re-implement CVFT and SAFA. Specifically, we keep the VGG16 as the backbone on both models unchanged and divide feature maps basing on Case \uppercase\expandafter{\romannumeral3}. For CVFT, we divide the final aligned feature maps into 8 parts and compute the loss of each corresponding part. For SAFA, the polar transform is retained. We first divide the feature maps into 8 parts, and then we use one SPE in SAFA to deal with each part separately before computing the loss. 
Combined with our partition strategy, CVFT+Ours boosts the R@1 accuracy from $61.43\%$ to $68.20\%$ (+$6.77\%$) and the R@Top1\% accuracy from $99.02\%$ to $99.30\%$ (+$0.28\%$). Similarly, SAFA+Ours can further improve the R@1 accuracy from $89.84\%$ to $92.83\%$ (+$2.99\%$) and the R@Top1\% accuracy from $99.64\%$ to $99.78\%$ (+$0.14\%$).

Significant performance improvements suggest that our method helps to mine more contextual information, yielding discriminative features.
\par
\textbf{Results on CVACT.} CVACT has a similar data pattern with CVUSA. Our method with ResNet-50~\cite{he2016deep} backbone achieves $79.99\%$ R@1, $90.63\%$ R@5, $92.56\%$ R@10 and $97.03\%$ R@Top1\% on CVACT. For a fair comparison, our method using VGG16 as the backbone also acquires competitive results. CVFT~\cite{shi_optimal_nodate}+Ours obtains the improvement with the R@1 accuracy from $61.05\%$ to $62.90\%$ (+$1.85\%$) and the R@Top1\% accuracy from $95.93\%$ to $97.22\%$ (+$1.29\%$). SAFA~\cite{shi_spatial-aware_nodate}+Ours increases the R@1 accuracy from $81.03\%$ to $83.66\%$ (+$2.63\%$) and the R@Top1\% accuracy from $98.17\%$ to $98.41\%$ (+$0.24\%$). The experimental results demonstrate that our method is still effective on CVACT.

\subsection{Ablation Studies}\label{ablation}
To verify the effectiveness of components in our model, we design several ablation studies.
\par
\begin{figure}[htbp]
  \centering
  \includegraphics[width=1\linewidth]{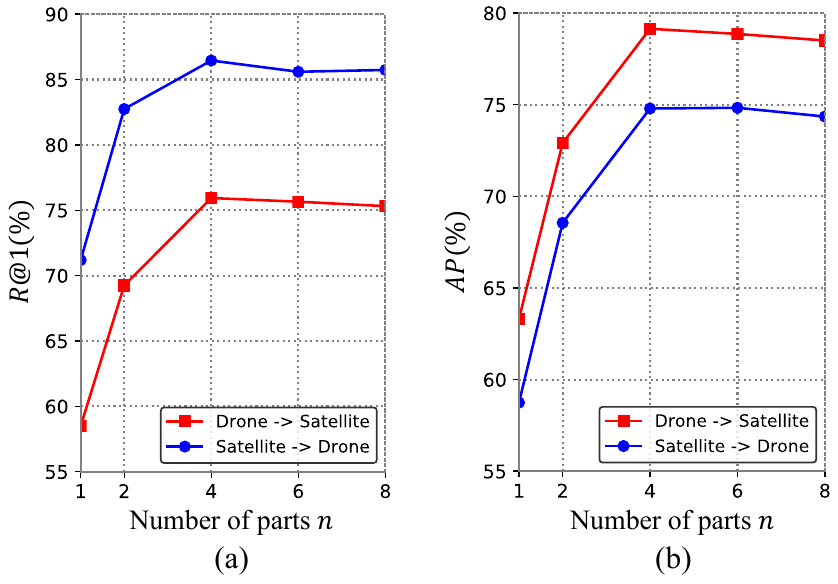}
  \caption{The effect of the number of parts $n$ on two tasks of the University-1652 dataset, \ie, Drone $\rightarrow$ Satellite and Satellite $\rightarrow$ Drone. The red line refers to the task of drone-view target localization (Drone $\rightarrow$ Satellite). The blue line shows the task of drone navigation (Satellite $\rightarrow$ Drone). We show the effect of the number of parts for R@1 accuracy (a), and AP accuracy (b). We observe that LPN achieves the best performance when the number of parts $n=$4.}
  \label{fig:parts}
\end{figure}

\textbf{Effect of the number of parts.} 
The number of parts $n$ is one of the key parameters in our network. By default, we deploy $n=4$. 
When $n$ = 1, the model is employed to global average pooling. At this time, the model equals the baseline with global attention ~\cite{zheng_university-1652_nodate}.
As shown in Figure \ref{fig:parts}, with the increment of $n$, both the Recall@1 and AP values have a significant improvement, since more contextual information has been taken into consideration. 
Intuitively, concatenating more contextual information parts can improve the discriminability of the final feature descriptor. However, we note that, as $n$ increases, each part contains less receptive fields with limited semantic information. 
As a result, a higher value of $n$ compromises the discriminability of the image representation. When $n$ = 6 or 8, the performance gains slowly or even slightly degrades. Therefore, we use $n$ = 4 as the default choice for our network, which balances the mining of the contextual information and the appropriate size of the receptive field. 

\setlength{\tabcolsep}{10pt}
\begin{table}[htp]
\small
\caption{Ablation study on the effect of different input sizes on University-1652. 
}
\begin{center}
\begin{tabular}{l|cc|cc}
\hline
\multirow{2}{*}{Image Size}& \multicolumn{2}{c|}{Drone $\rightarrow$ Satellite} & \multicolumn{2}{c}{Satellite $\rightarrow$ Drone}\\
  & R@1 & AP & R@1 & AP\\
\shline
224 & 69.28 & 72.98 & 82.45 & 68.92 \\
256 & 75.93 & 79.14 & 86.45 & 74.79 \\
320 & 77.65 & 80.56 & 85.31 & 75.36 \\
384 & 78.02 & 80.99 & 86.16 & 76.56 \\
512 & 77.71 & 80.80 & 90.30 & 78.78 \\
\hline
\end{tabular}
\end{center}
\label{table:Size}
\end{table}

\textbf{Effect of the input image size.}
A small training size compresses the fine-grained information of the input image, which compromises the discriminative representation learning. 
In contrast, a larger input size introduces more memory costs during training. 
To balance the input image size with the memory usage, we study the effect of the input image size. We just change the image size and the covered region of the image is not changed in the experiment. As shown in Table \ref{table:Size}, in both tasks, \ie, (Drone $\rightarrow$ Satellite) and (Satellite $\rightarrow$ Drone), as the input image size from 224 to 384, we observe that the performance gradually improves. 
When we continue to enlarge the input size to 512, the improvement is not clear on the Drone $\rightarrow$ Satellite task. We hope this study could help the real-world application in selecting the appropriate input size, when computation sources are limited.  

\setlength{\tabcolsep}{8pt}
\begin{table}
\small
\caption{Ablation study on rotating images during inference on University-1652. 
}
\begin{center}
\begin{tabular}{ll|cc|cc}
\hline
\multicolumn{2}{c|}{Rotation Angle}& \multicolumn{2}{c|}{Drone $\rightarrow$ Satellite} & \multicolumn{2}{c}{Satellite $\rightarrow$ Drone}\\
  Query & Gallery & R@1 & AP & R@1 & AP\\
\shline
$0^\circ$ & $0^\circ$ & 75.93 & 79.14 & 86.45 & 74.79 \\
$16^\circ$ & $0^\circ$ & 75.64 & 78.86 & 85.16 & 72.78 \\
$45^\circ$ & $0^\circ$ & 72.04 & 75.62 & 85.16 & 72.27 \\
$67^\circ$ & $0^\circ$ & 70.39 & 74.09 & 85.73 & 73.06 \\
$90^\circ$ & $0^\circ$ & 68.80 & 72.67 & 86.31 & 75.31 \\
$180^\circ$ & $0^\circ$ & 70.76 & 74.47 & 85.45 & 74.03 \\
$204^\circ$ & $0^\circ$ & 69.92 & 73.68 & 84.45 & 72.22 \\
$270^\circ$ & $0^\circ$ & 69.06 & 72.49 & 86.73 & 75.12 \\
$317^\circ$ & $0^\circ$ & 72.29 & 75.87 & 84.17 & 71.85 \\
$32^\circ$ & $75^\circ$ & 73.19 & 76.69 & 83.17 & 66.41 \\
$216^\circ$ & $87^\circ$ & 69.54 & 73.27 & 83.45 & 65.29 \\
\hline
\end{tabular}
\end{center}
\label{table:Rotation}
\end{table}

\textbf{Is LPN robust to rotation variants?}
Satellite-view images in University-1652 are north aligned and the orientation of drone-view images is random. In the training phase, the rotation augmentation is applied in the satellite-view branch but not in other branches. To verify the scalability of LPN for image rotation, we conduct experiments on rotating the query image to retrieval the true-matched images. We do not rotate gallery images but query images. The experimental results are shown in Table \ref{table:Rotation}. The $0^\circ$ denotes the input query image without rotation. 
For the task of drone navigation (Satellite $\rightarrow$ Drone), LPN obtains robust features for unseen satellite-view query images against different rotation angles. In contrast, for the task of drone-view target localization (Drone $\rightarrow$ Satellite), we do not train the model on the rotated drone-view data. The LPN still achieves one competitive performance without a significant performance drop. In addition, we also attempt to rotate different angles on query and gallery images to further test our model. The experimental results suggest that LPN has good scalability to rotation variations.

\begin{figure}[t]
  \centering
  \includegraphics[width=1\linewidth]{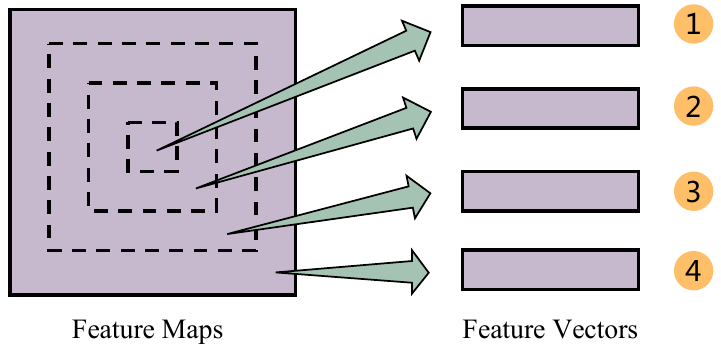}
  \caption{The feature maps are first divided into four parts in LPN. Then, an average pooling layer transforms these four parts into four column vectors which are treated as subsequent feature descriptors. For each part, we use the numbers $1, 2, 3, 4$ to represent. }
  \label{fig:each_part}
\end{figure}

\setlength{\tabcolsep}{1pt}
\begin{table}[t]
\small
\caption{Ablation study of using one part or a combination of different parts during inference. $1$, $2$, $3$, $4$ indicate four averaged parts which are sliced from feature maps.
}
\begin{center}
\begin{tabular}{ll|cc|cc}
\hline
\multicolumn{2}{c|}{Part Combination}& \multicolumn{2}{c|}{Drone $\rightarrow$ Satellite} & \multicolumn{2}{c}{Satellite $\rightarrow$ Drone}\\
 Query & Gallery & R@1 & AP & R@1 & AP\\
\shline
$1$ & $1$ & 71.95 & 75.50 & 84.02 & 70.24 \\
$2$ & $2$ & 71.94 & 75.49 & 83.74 & 71.32 \\
$3$ & $3$ & 71.97 & 75.62 & 85.45 & 71.61 \\
$4$ & $4$ & 70.75 & 74.41 & 82.03 & 69.50 \\
$1+2$ & $1+2$ & 74.85 & 78.14 & 85.59 & 73.69 \\
$1+2+3$ & $1+2+3$ & 75.74 & 78.97 & 86.31 & 74.76 \\
$1+2+3+4$ & $1+2+3+4$ & \textbf{75.93} & \textbf{79.14} & \textbf{86.45} & \textbf{74.79} \\
$1+2+3$ & $2+3+4$ & 0.07 & 0.38 & 0.14 & 0.21 \\
$1+2$ & $2+3$ & 0.08 & 0.39 & 0.00 & 0.17 \\
$1+2$ & $3+4$ & 0.13 & 0.50 & 0.00 & 0.21 \\
\hline
\end{tabular}
\end{center}
\label{table:part_influence}
\end{table}

\begin{figure}[t]
  \centering
  \includegraphics[width=1\linewidth]{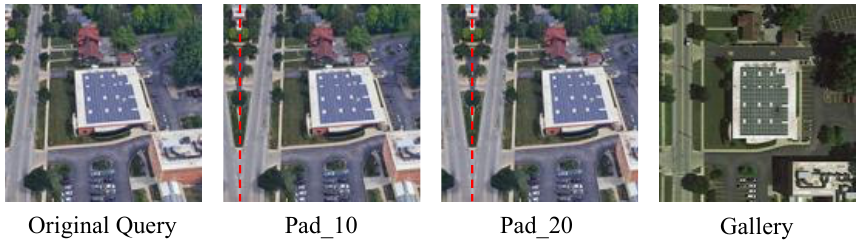}
  \caption{The first and last images are the original query and gallery images in the test set. We pad 10 and 20 pixels in the left of the query image in the way of reflection, respectively. Then, we crop the padded image to the original image size in a left-aligned manner. Thus we can obtain the second and third images, which have an offset of the geographic target with the gallery image. The left space of the red dotted line is the extra padded pixels.}
  \label{fig:pad}
\end{figure}

\setlength{\tabcolsep}{10pt}
\begin{table}
\small
\caption{Ablation study on shifting query images during inference on University-1652.
}
\begin{center}
\begin{tabular}{l|cc|cc}
\hline
\multirow{2}{*}{Shifted Pixel}& \multicolumn{2}{c|}{Drone $\rightarrow$ Satellite} & \multicolumn{2}{c}{Satellite $\rightarrow$ Drone}\\
  & R@1 & AP & R@1 & AP\\
\shline
0 & 75.93 & 79.14 & 86.45 & 74.79 \\
10 & 75.26 & 78.57 & 85.16 & 72.84 \\
20 & 72.37 & 76.04 & 83.02 & 70.67 \\
\hline
\end{tabular}
\end{center}
\label{table:shift}
\end{table}


\textbf{Does LPN learn complementary part features?}
The square-ring partition strategy divides the feature maps into four parts in LPN. We use the numbers 1, 2, 3, and 4 to represent the four parts of the feature maps, as shown in Figure~\ref{fig:each_part}. Subsequently, we conduct experiments by choosing one or a combination of the four parts. The experimental results demonstrate LPN has learned complementary features (see Table~\ref{table:part_influence}). We observe that using only one part has obtained one fairly good result in two tasks, \ie, (Drone $\rightarrow$ Satellite) and (Satellite $\rightarrow$ Drone). When we further concatenate two or three parts, the accuracy of Recall@1 and AP gradually increases. 
When all parts are leveraged, we obtain the best performance in both tasks. It demonstrates that LPN has learned complementary part feature, enriching the model capability. But the Recall@1 accuracy and AP drop dramatically in both tasks if query features and gallery features cover different parts. The results reflect from the side that the learned semantic information between the parts is complementary. Because of the complementary in each part, the semantic information contained in different parts is not overlapping. When there are different parts in the query and gallery feature (\ie, the true-matched image pair covers significant different areas), the different parts become the distractors for the final visual feature, resulting in terrible retrieval performance. 
\par
\textbf{Is LPN robust to the shifted query image?} In the realistic scenario, there is usually an offset in the geographic target location of the query image and true-matched images in the gallery. To explore whether LPN can cope with the offset of the geographic target location in a true-matched image pair, we carry out experiments on shifting the query image during testing. Specifically, we shift the query image to the right in pixels and keep images in the gallery intact (see Figure \ref{fig:pad}). Table \ref{table:shift} shows the experimental results. 0 indicates that the input query image is not offset. When the input query image is shifted 10 pixels, we can hardly observe a performance drops for drone navigation and drone-view target localization tasks. While the shifted pixels is 20, the performance on both tasks decreases slightly. The experimental results suggest that LPN is robust when there is a small offset of the geographic target location for a true-matched image pair in retrieval.
\setlength{\tabcolsep}{6pt}
\begin{table}
\small
\caption{The performance of geo-localization between satellite-view images and ground-view images. (w/o drone) indicates that LPN is trained without drone-view images.
}
\begin{center}
\begin{tabular}{l|cc|cc}
\hline
\multirow{2}{*}{Method}& \multicolumn{2}{c|}{Satellite $\rightarrow$ Ground} & \multicolumn{2}{c}{Ground $\rightarrow$ Satellite}\\
  & R@1 & AP & R@1 & AP\\
\shline
Baseline~\cite{zheng_university-1652_nodate} & 1.14 & 1.14 & 1.20 & 2.52\\
Ours (w/o drone) & 1.43 & 1.31 & 0.74 & 1.83 \\
Ours & 1.85 & 1.66 & 0.81 & 2.21 \\
\hline
\end{tabular}
\end{center}
\label{table:ground}
\end{table}

\textbf{Geo-localization between satellite-view images and ground-view images.} In University-1652, geo-localization between satellite-view images and ground-view images is a challenging task. We can sum up the difficulties in the following points. $(1)$ Ground and satellite images that are collected from different viewpoints naturally have a distinct visual appearance. $(2)$ Unlike ground-view images in CVUSA and CVACT, which are panoramas, the ground-view image in University-1652 only covers part of the whole building. $(3)$ Ground-view images in University-1652 contains vast obstacles, \eg, trees and cars. From Table \ref{table:ground}, we can notice that geo-localization between satellite-view and ground-view images do not work well. Using satellite-view images to retrieve the ground-view images, our method achieves the best results in Recall@1 accuracy and AP. Whether employing drone-view images or not, there is a performance decrease in the ground-to-satellite localization compared with baseline~\cite{zheng_university-1652_nodate}. But using drone-view images in LPN obtains better results than without using these.
\par
\textbf{Effect of the drone distance to the geographic target.} The scale of the satellite-view image in University-1652 is fixed, while the scale of the drone-view image changes dynamically with the drone distance to the geographic target. We adopt drone-view images with different distances to the geographic target as queries to study the impact of the changed scale for LPN. As shown in Table \ref{table:scale}, when the drone-view image is captured in a middle distance to the geographic target, we obtain the best performance. When the drone distance is short to the geographic target, we can observe that the results are still competitive compared with using all drone-view query images. The scales of these drone-view images are close to satellite-view images. Another reason is that these drone-view images mainly contain the target building without extra trees and other buildings.
\par
\setlength{\tabcolsep}{28pt}
\begin{table}
\small
\caption{Ablation study on using drone-view images with different distance to the geographic target to conduct retrieval. "All" indicates that we apply all drone-view query images.
}
\begin{center}
\begin{tabular}{l|cc}
\hline
\multirow{2}{*}{Distance}& \multicolumn{2}{c}{Drone $\rightarrow$ Satellite} \\
  & R@1 & AP \\
\shline
All  & 75.93 & 79.14  \\
Long & 60.20 & 65.01 \\
Short  & 75.04 & 78.35  \\
Middle & 80.03 & 82.77  \\
\hline
\end{tabular}
\end{center}
\label{table:scale}
\end{table}

\begin{figure}[htp]
  \centering
  \includegraphics[width=1\linewidth]{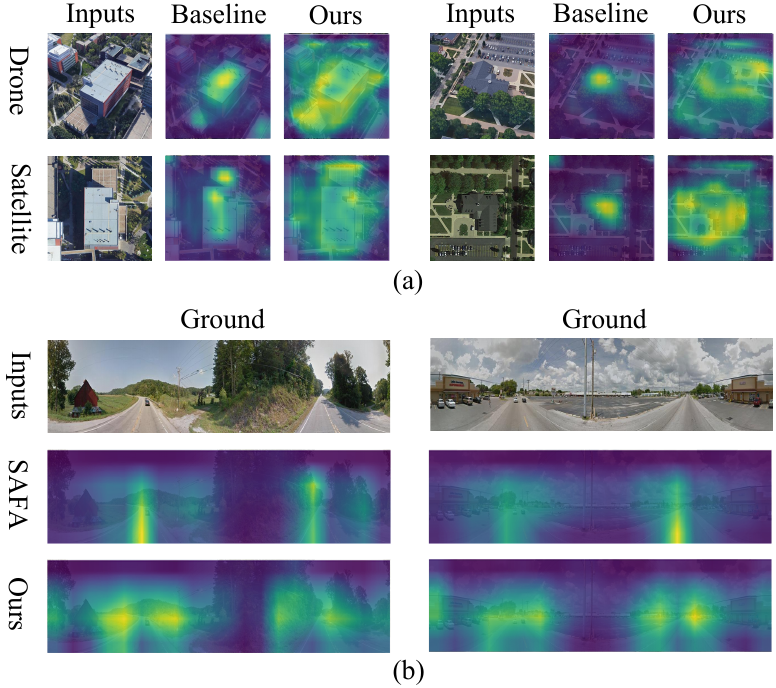}
  \caption{
  Visualization of heatmaps. (a) Heatmaps generated by baseline~\cite{zheng_university-1652_nodate} and ours in different platforms of University-1652. (b) Ground-view heatmaps learned from SAFA~\cite{shi_spatial-aware_nodate} and ours (SAFA + ours) on CVUSA. 
  }
  \label{fig:visual_heatmap}
\end{figure}
\textbf{Transfer learning from University-1652 to small-scale datasets.} 
To study the generalization ability of LPN trained on University-1652, we evaluate three models on two small-scale landmark retrieval datasets, \ie, Oxford5k~\cite{philbin2007object} and Paris6k~\cite{philbin2008lost}. The first model is ResNet-50~\cite{he2016deep} trained on ImageNet~\cite{5206848}. The second model is baseline~\cite{zheng_university-1652_nodate} and LPN is the third model. During the evaluation, three models have not been fine-tuned on these two datasets. For baseline and LPN, we choose two different branches, \ie, $\mathcal{F}_s$ and $\mathcal{F}_g$ to extract features, since these two branches focus on different low-level patterns of input images. Weights on $\mathcal{F}_s$ are trained by the satellite-view images, while $\mathcal{F}_g$ is learned on the ground-view images. From Table~\ref{table:transfer}, we observe that the extracted feature from LPN shows better performance on both two datasets than features obtained from ResNet-50 and baseline. This result also demonstrates that the square-ring partition strategy can enhance the generalization ability of our model.
We also note that in the same model, $\mathcal{F}_g$ has a superior generalization ability than $\mathcal{F}_s$. 
It is because that images in Oxford5k and Pairs6k are closer to the Google Street View images, which are similar to the ground-view images collected from Google Image. Besides, $\mathcal{F}_s$ is trained by the aerial-view data, which viewpoint is perpendicular to the ground plane. In contrast, the data viewpoint in Oxford5k or Paris6k is parallel to the ground plane.

\setlength{\tabcolsep}{6pt}
\begin{table}
\small
\caption{Transfer learning from University-1652 to small-scale datasets, \ie, Oxford5k~\cite{philbin2007object} and Paris6k~\cite{philbin2008lost}. We show the AP ($\%$) accuracy on two datasets.
}
\begin{center}
\begin{tabular}{l|c|cc|cc}
\hline
\multirow{2}{*}{Dataset} & \multirow{2}{*}{ResNet-50} & \multicolumn{2}{c|}{baseline~\cite{zheng_university-1652_nodate}} & \multicolumn{2}{c}{Ours}\\
\cline{3-6}  & & $\mathcal{F}_s$ & $\mathcal{F}_g$ & $\mathcal{F}_s$ & $\mathcal{F}_g$\\
\shline
Oxford5k~\cite{philbin2007object} & 8.43 & 15.62 & 41.12 & 27.02 & 51.71 \\  
Paris6k~\cite{philbin2008lost} & 27.93 & 38.18 & 59.00 & 45.81 & 67.73 \\ 

\hline
\end{tabular}
\end{center}
\label{table:transfer}
\end{table}

\begin{figure}[t]
  \centering
  \includegraphics[width=1\linewidth]{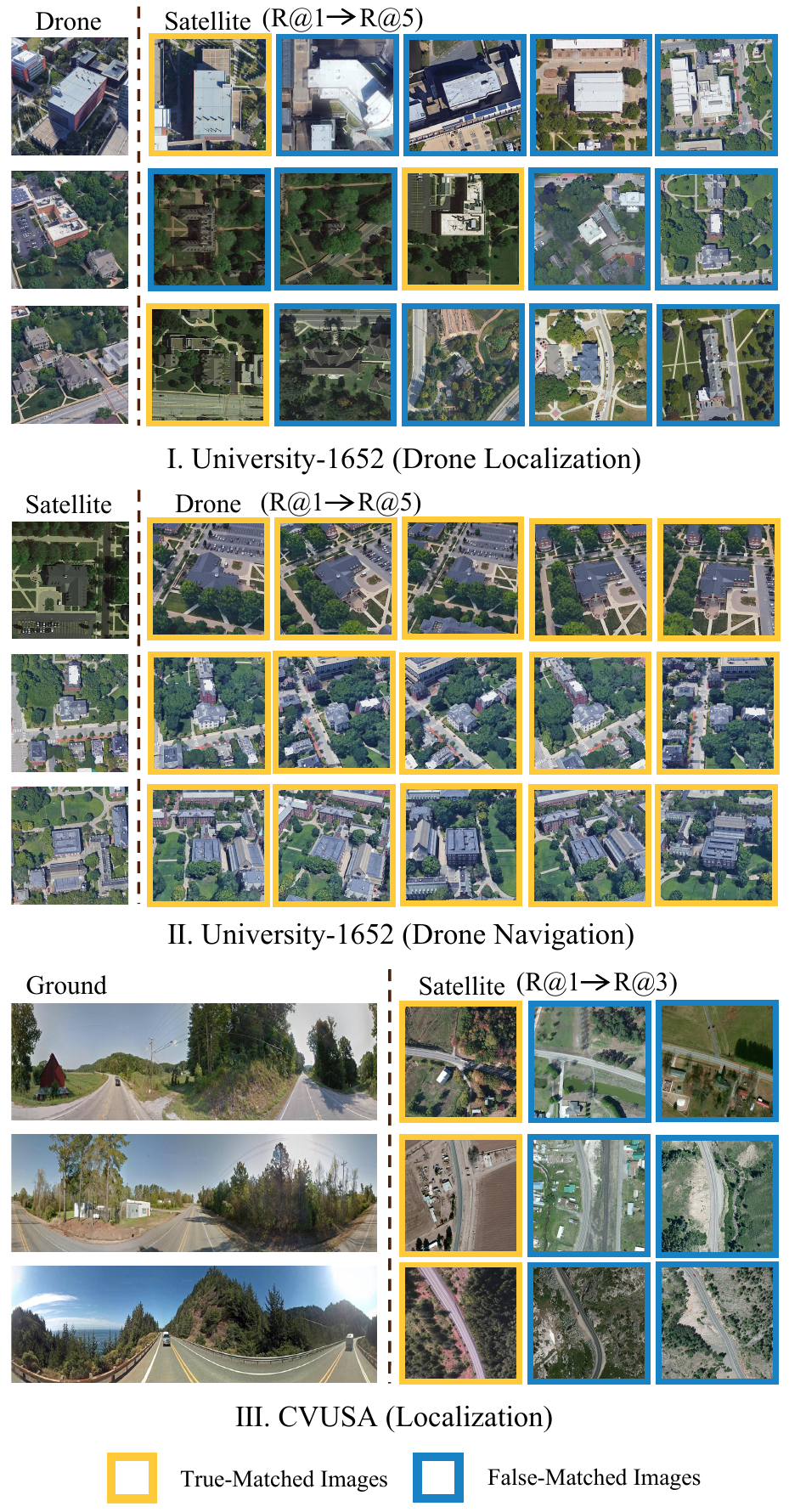}
  \caption{
  Qualitative image retrieval results. (\uppercase\expandafter{\romannumeral1}) Top-5 retrieval results of drone-view target localization on University-1652. (\uppercase\expandafter{\romannumeral2}) Top-5 retrieval results of drone navigation on University-1652. (\uppercase\expandafter{\romannumeral3}) Top-3 retrieval results of geographic localization on CVUSA. The true matches are in yellow boxes, while the false matches are displayed in blue boxes.
  }
  \label{fig:visual1}
\end{figure}

\subsection{Qualitative Result} 
As an additional qualitative evaluation, we visualize some heatmaps created by our and compared methods. Figure \ref{fig:visual_heatmap} (a) shows heatmaps generated by baseline~\cite{zheng_university-1652_nodate} and LPN in the drone and satellite platforms. Compared with the baseline, our approach activates the region of the geographic target and neighbor areas containing the contextual information. Figure \ref{fig:visual_heatmap} (b) shows the ground-view heatmaps generated by the original SAFA~\cite{shi_spatial-aware_nodate} and SAFA fused our partition strategy (Ours). SAFA only activates the position of the road, while our method further places emphasis on contextual information next to the road position. Our method is more consistent with the processing of the human visual system to locate an unfamiliar road.

Moreover, we show some retrieval results for different tasks on University-1652 and CVUSA in Figure \ref{fig:visual1}. On University-1652, we observe that LPN can adapt to retrieve the reasonable images based on the content in both drone-view localization and drone navigation tasks. One failure case is also shown in the second row of Figure \ref{fig:visual1} (\uppercase\expandafter{\romannumeral1}), in which LPN can not recall the matched image in top-1. We notice that it is challenging in that the recalled top-1 image has a very similar pattern with the query image, especially the appearance of the geographic target in two images. 
On CVUSA, we observe a similar result. Our method with SAFA~\cite{shi_spatial-aware_nodate} has successfully retrieved the relevant satellite-view images.

\section{Conclusion}\label{conclusion}
In this paper, we identify the challenge in cross-view geo-localization, and propose a simple and effective deep neural network, called Local Pattern Network (LPN), to explicitly mine the contextual information. Specifically, we introduce a square-ring partition strategy for learning complementary spatial features according to the distance to the image center. The contextual information enhances the discriminability of the image representation with more fine-grained patterns. Our approach has achieved competitive accuracy on three cross-view geo-localization benchmarks, \ie, University-1652, CVUSA and CVACT. Moreover, the proposed LPN has good scalability to rotation variation, which is close to the real-world application. The square-ring partition strategy also can be easily embedded into other frameworks to boost performance. In the future, we will investigate applying a module, such as STN~\cite{jaderberg_spatial_2015}, to estimate the scale of the drone-view image.

\ifCLASSOPTIONcaptionsoff
  \newpage
\fi


\bibliographystyle{IEEEtran}
\bibliography{IEEEabrv,partmatter}

\begin{thebibliography}{10}
\providecommand{\url}[1]{#1}
\csname url@samestyle\endcsname
\providecommand{\newblock}{\relax}
\providecommand{\bibinfo}[2]{#2}
\providecommand{\BIBentrySTDinterwordspacing}{\spaceskip=0pt\relax}
\providecommand{\BIBentryALTinterwordstretchfactor}{4}
\providecommand{\BIBentryALTinterwordspacing}{\spaceskip=\fontdimen2\font plus
\BIBentryALTinterwordstretchfactor\fontdimen3\font minus
  \fontdimen4\font\relax}
\providecommand{\BIBforeignlanguage}[2]{{%
\expandafter\ifx\csname l@#1\endcsname\relax
\typeout{** WARNING: IEEEtran.bst: No hyphenation pattern has been}%
\typeout{** loaded for the language `#1'. Using the pattern for}%
\typeout{** the default language instead.}%
\else
\language=\csname l@#1\endcsname
\fi
#2}}
\providecommand{\BIBdecl}{\relax}
\BIBdecl

\bibitem{shi_spatial-aware_nodate}
Y.~Shi, L.~Liu, X.~Yu, and H.~Li, ``Spatial-aware feature aggregation for image
  based cross-view geo-localization,'' in \emph{Neural Information Processing
  Systems}, 2019.

\bibitem{zheng_university-1652_nodate}
Z.~Zheng, Y.~Wei, and Y.~Yang, ``University-1652: A multi-view multi-source
  benchmark for drone-based geo-localization,'' in \emph{ACM International
  Conference on Multimedia}, 2020, doi:{
  \href{http://dx.doi.org/10.1145/3394171.3413896}{10.1145/3394171.3413896}}.

\bibitem{workman_location_2015}
S.~Workman and N.~Jacobs, ``On the location dependence of convolutional neural
  network features,'' in \emph{IEEE Conference on Computer Vision and Pattern
  Recognition}, 2015.

\bibitem{liu_lending_2019}
L.~Liu and H.~Li, ``Lending orientation to neural networks for cross-view
  geo-localization,'' in \emph{IEEE Conference on Computer Vision and Pattern
  Recognition}, 2019.

\bibitem{he2016deep}
K.~He, X.~Zhang, S.~Ren, and J.~Sun, ``Deep residual learning for image
  recognition,'' in \emph{IEEE Conference on Computer Vision and Pattern
  Recognition}, 2016.

\bibitem{zheng2020dual}
Z.~Zheng, L.~Zheng, M.~Garrett, Y.~Yang, M.~Xu, and Y.-D. Shen, ``Dual-path
  convolutional image-text embeddings with instance loss,'' \emph{ACM
  Transactions on Multimedia Computing, Communications, and Applications
  (TOMM)}, vol.~16, no.~2, pp. 1--23, 2020, doi:{
  \href{http://dx.doi.org/10.1145/3383184}{10.1145/3383184}}.

\bibitem{shi_optimal_nodate}
Y.~Shi, X.~Yu, L.~Liu, T.~Zhang, and H.~Li, ``Optimal feature transport for
  cross-view image geo-localization,'' in \emph{AAAI Conference on Artificial
  Intelligence}, 2020.

\bibitem{Shi_2020_CVPR}
Y.~Shi, X.~Yu, D.~Campbell, and H.~Li, ``Where am i looking at? joint location
  and orientation estimation by cross-view matching,'' in \emph{IEEE Conference
  on Computer Vision and Pattern Recognition}, 2020.

\bibitem{hu_cvm-net_2018}
S.~Hu, M.~Feng, R.~M. Nguyen, and G.~Hee~Lee, ``Cvm-net: Cross-view matching
  network for image-based ground-to-aerial geo-localization,'' in \emph{IEEE
  Conference on Computer Vision and Pattern Recognition}, 2018.

\bibitem{hadsell2006dimensionality}
R.~Hadsell, S.~Chopra, and Y.~LeCun, ``Dimensionality reduction by learning an
  invariant mapping,'' in \emph{IEEE Conference on Computer Vision and Pattern
  Recognition}, 2006.

\bibitem{deng2018image}
W.~Deng, L.~Zheng, Q.~Ye, G.~Kang, Y.~Yang, and J.~Jiao, ``Image-image domain
  adaptation with preserved self-similarity and domain-dissimilarity for person
  re-identification,'' in \emph{IEEE Conference on Computer Vision and Pattern
  Recognition}, 2018.

\bibitem{schroff2015facenet}
F.~Schroff, D.~Kalenichenko, and J.~Philbin, ``Facenet: A unified embedding for
  face recognition and clustering,'' in \emph{IEEE Conference on Computer
  Vision and Pattern Recognition}, 2015.

\bibitem{fu2019sta}
Y.~Fu, X.~Wang, Y.~Wei, and T.~Huang, ``Sta: Spatial-temporal attention for
  large-scale video-based person re-identification,'' in \emph{AAAI Conference
  on Artificial Intelligence}, 2019.

\bibitem{rensink2000dynamic}
R.~A. Rensink, ``The dynamic representation of scenes,'' \emph{Visual
  Cognition}, vol.~7, no. 1-3, pp. 17--42, 2000.

\bibitem{corbetta2002control}
M.~Corbetta and G.~L. Shulman, ``Control of goal-directed and stimulus-driven
  attention in the brain,'' \emph{Nature Reviews Neuroscience}, vol.~3, no.~3,
  pp. 201--215, 2002.

\bibitem{zheng2020vehiclenet}
Z.~Zheng, T.~Ruan, Y.~W. Wei, Y.~Yang, and M.~Tao, ``Vehiclenet: Learning
  robust visual representation for vehicle re-identification,'' \emph{IEEE
  Transactions on Multimedia (TMM)}, 2020, doi:{
  \href{http://dx.doi.org/10.1109/TMM.2020.3014488}{10.1109/TMM.2020.3014488}}.

\bibitem{zhai_predicting_2017}
M.~Zhai, Z.~Bessinger, S.~Workman, and N.~Jacobs, ``Predicting ground-level
  scene layout from aerial imagery,'' in \emph{IEEE Conference on Computer
  Vision and Pattern Recognition}, 2017.

\bibitem{SemanticCM}
F.~Castaldo, A.~R. Zamir, R.~Angst, F.~A.~N. Palmieri, and S.~Savarese,
  ``Semantic cross-view matching,'' in \emph{IEEE International Conference on
  Computer Vision Workshops}, 2015.

\bibitem{lin2013cross}
T.-Y. Lin, S.~Belongie, and J.~Hays, ``Cross-view image geolocalization,'' in
  \emph{IEEE Conference on Computer Vision and Pattern Recognition}, 2013.

\bibitem{senlet2011framework}
T.~Senlet and A.~Elgammal, ``A framework for global vehicle localization using
  stereo images and satellite and road maps,'' in \emph{IEEE International
  Conference on Computer Vision Workshops}, 2011.

\bibitem{bansal2011geo}
M.~Bansal, H.~S. Sawhney, H.~Cheng, and K.~Daniilidis, ``Geo-localization of
  street views with aerial image databases,'' in \emph{ACM International
  Conference on Multimedia}, 2011.

\bibitem{workman_wide-area_2015}
S.~Workman, R.~Souvenir, and N.~Jacobs, ``Wide-area image geolocalization with
  aerial reference imagery,'' in \emph{IEEE International Conference on
  Computer Vision}, 2015.

\bibitem{lin_learning_2015}
T.-Y. Lin, Y.~Cui, S.~Belongie, and J.~Hays, ``Learning deep representations
  for ground-to-aerial geolocalization,'' in \emph{IEEE Conference on Computer
  Vision and Pattern Recognition}, 2015.

\bibitem{chopra2005learning}
S.~Chopra, R.~Hadsell, and Y.~LeCun, ``Learning a similarity metric
  discriminatively, with application to face verification,'' in \emph{IEEE
  Conference on Computer Vision and Pattern Recognition}, 2005.

\bibitem{arandjelovic2016netvlad}
R.~Arandjelovic, P.~Gronat, A.~Torii, T.~Pajdla, and J.~Sivic, ``Netvlad: Cnn
  architecture for weakly supervised place recognition,'' in \emph{IEEE
  Conference on Computer Vision and Pattern Recognition}, 2016.

\bibitem{vo_localizing_2017}
N.~N. Vo and J.~Hays, ``Localizing and orienting street views using overhead
  imagery,'' in \emph{European Conference on Computer Vision}, 2016.

\bibitem{li2020hierarchical}
P.~Li, P.~Pan, P.~Liu, M.~Xu, and Y.~Yang, ``Hierarchical temporal modeling
  with mutual distance matching for video based person re-identification,''
  \emph{IEEE Transactions on Circuits and Systems for Video Technology}, 2020.

\bibitem{zheng2017unlabeled}
Z.~Zheng, L.~Zheng, and Y.~Yang, ``Unlabeled samples generated by gan improve
  the person re-identification baseline in vitro,'' in \emph{IEEE International
  Conference on Computer Vision}, 2017, doi:{
  \href{http://dx.doi.org/10.1109/ICCV.2017.405}{10.1109/ICCV.2017.405}}.

\bibitem{amit2007pop}
Y.~Amit and A.~Trouv{\'e}, ``Pop: Patchwork of parts models for object
  recognition,'' \emph{International Journal of Computer Vision}, vol.~75,
  no.~2, pp. 267--282, 2007.

\bibitem{crandall2005spatial}
D.~Crandall, P.~Felzenszwalb, and D.~Huttenlocher, ``Spatial priors for
  part-based recognition using statistical models,'' in \emph{IEEE Conference
  on Computer Vision and Pattern Recognition}, 2005.

\bibitem{fergus2003object}
R.~Fergus, P.~Perona, and A.~Zisserman, ``Object class recognition by
  unsupervised scale-invariant learning,'' in \emph{IEEE Conference on Computer
  Vision and Pattern Recognition}, 2003.

\bibitem{leibe2008robust}
B.~Leibe, A.~Leonardis, and B.~Schiele, ``Robust object detection with
  interleaved categorization and segmentation,'' \emph{International Journal of
  Computer Vision}, vol.~77, no. 1-3, pp. 259--289, 2008.

\bibitem{weber2000towards}
M.~Weber, M.~Welling, and P.~Perona, ``Towards automatic discovery of object
  categories,'' in \emph{IEEE Conference on Computer Vision and Pattern
  Recognition}, 2000.

\bibitem{LBP}
T.~{Ojala}, M.~{Pietikainen}, and T.~{Maenpaa}, ``Multiresolution gray-scale
  and rotation invariant texture classification with local binary patterns,''
  \emph{IEEE Transactions on Pattern Analysis and Machine Intelligence},
  vol.~24, no.~7, pp. 971--987, 2002.

\bibitem{SIFT}
D.~G. Lowe, ``Object recognition from local scale-invariant features,'' in
  \emph{IEEE International Conference on Computer Vision}, 1999.

\bibitem{zhao_spindle_2017}
H.~Zhao, M.~Tian, S.~Sun, J.~Shao, J.~Yan, S.~Yi, X.~Wang, and X.~Tang,
  ``Spindle net: Person re-identification with human body region guided feature
  decomposition and fusion,'' in \emph{IEEE Conference on Computer Vision and
  Pattern Recognition}, 2017.

\bibitem{xu_attention-aware_2018}
J.~Xu, R.~Zhao, F.~Zhu, H.~Wang, and W.~Ouyang, ``Attention-aware compositional
  network for person re-identification,'' in \emph{IEEE Conference on Computer
  Vision and Pattern Recognition}, 2018.

\bibitem{guo_beyond_2019}
J.~Guo, Y.~Yuan, L.~Huang, C.~Zhang, J.-G. Yao, and K.~Han, ``Beyond human
  parts: Dual part-aligned representations for person re-identification,'' in
  \emph{IEEE International Conference on Computer Vision}, 2019.

\bibitem{li_learning_2017}
D.~Li, X.~Chen, Z.~Zhang, and K.~Huang, ``Learning deep context-aware features
  over body and latent parts for person re-identification,'' in \emph{IEEE
  Conference on Computer Vision and Pattern Recognition}, 2017.

\bibitem{zheng2018pedestrian}
Z.~Zheng, L.~Zheng, and Y.~Yang, ``Pedestrian alignment network for large-scale
  person re-identification,'' \emph{IEEE Transactions on Circuits and Systems
  for Video Technology}, vol.~29, no.~10, pp. 3037--3045, 2018, doi:{
  \href{http://dx.doi.org/10.1109/TCSVT.2018.2873599}{10.1109/TCSVT.2018.2873599}}.

\bibitem{jaderberg_spatial_2015}
M.~Jaderberg, K.~Simonyan, A.~Zisserman \emph{et~al.}, ``Spatial transformer
  networks,'' in \emph{Neural Information Processing Systems}, 2015.

\bibitem{zhao_deeply-learned_2017}
L.~Zhao, X.~Li, Y.~Zhuang, and J.~Wang, ``Deeply-learned part-aligned
  representations for person re-identification,'' in \emph{IEEE International
  Conference on Computer Vision}, 2017.

\bibitem{sun_beyond_2018}
Y.~Sun, L.~Zheng, Y.~Yang, Q.~Tian, and S.~Wang, ``Beyond part models: Person
  retrieval with refined part pooling (and a strong convolutional baseline),''
  in \emph{European Conference on Computer Vision}, 2018.

\bibitem{sun_dissecting_2019}
X.~Sun and L.~Zheng, ``Dissecting person re-identification from the viewpoint
  of viewpoint,'' in \emph{IEEE Conference on Computer Vision and Pattern
  Recognition}, 2019.

\bibitem{zhong_invariance_2019}
Z.~Zhong, L.~Zheng, Z.~Luo, S.~Li, and Y.~Yang, ``Invariance matters: Exemplar
  memory for domain adaptive person re-identification,'' in \emph{IEEE
  Conference on Computer Vision and Pattern Recognition}, 2019.

\bibitem{song_generalizable_2019}
J.~Song, Y.~Yang, Y.-Z. Song, T.~Xiang, and T.~M. Hospedales, ``Generalizable
  person re-identification by domain-invariant mapping network,'' in \emph{IEEE
  Conference on Computer Vision and Pattern Recognition}, 2019.

\bibitem{fu2019horizontal}
Y.~Fu, Y.~Wei, Y.~Zhou, H.~Shi, G.~Huang, X.~Wang, Z.~Yao, and T.~Huang,
  ``Horizontal pyramid matching for person re-identification,'' in \emph{AAAI
  Conference on Artificial Intelligence}, 2019.

\bibitem{zheng_pose-invariant_2019}
L.~Zheng, Y.~Huang, H.~Lu, and Y.~Yang, ``Pose-invariant embedding for deep
  person re-identification,'' \emph{IEEE Transactions on Image Processing},
  vol.~28, no.~9, pp. 4500--4509, 2019.

\bibitem{su_pose-driven_2017}
C.~Su, J.~Li, S.~Zhang, J.~Xing, W.~Gao, and Q.~Tian, ``Pose-driven deep
  convolutional model for person re-identification,'' in \emph{IEEE
  International Conference on Computer Vision}, 2017.

\bibitem{wei_glad_2019}
L.~Wei, S.~Zhang, H.~Yao, W.~Gao, and Q.~Tian, ``Glad: Global--local-alignment
  descriptor for scalable person re-identification,'' \emph{IEEE Transactions
  on Multimedia}, vol.~21, no.~4, pp. 986--999, 2018.

\bibitem{vgg}
K.~Simonyan and A.~Zisserman, ``Very deep convolutional networks for
  large-scale image recognition,'' in \emph{ICLR}, 2015.

\bibitem{philbin2007object}
J.~Philbin, O.~Chum, M.~Isard, J.~Sivic, and A.~Zisserman, ``Object retrieval
  with large vocabularies and fast spatial matching,'' in \emph{IEEE Conference
  on Computer Vision and Pattern Recognition}, 2007.

\bibitem{philbin2008lost}
J.~Philbin, O.~Chum, M.~Isard, J.~Sivic, and A.~Zisserman, ``Lost in
  quantization: Improving particular object retrieval in large scale image
  databases,'' in \emph{IEEE Conference on Computer Vision and Pattern
  Recognition}, 2008.

\bibitem{5206848}
J.~Deng, W.~Dong, R.~Socher, L.~J. Li, K.~Li, and L.~Fei-Fei, ``Imagenet: A
  large-scale hierarchical image database,'' in \emph{IEEE Conference on
  Computer Vision and Pattern Recognition}, 2009.

\bibitem{kaiming_init}
K.~He, X.~Zhang, S.~Ren, and J.~Sun, ``Delving deep into rectifiers: Surpassing
  human-level performance on imagenet classification,'' in \emph{IEEE
  International Conference on Computer Vision}, 2015.

\bibitem{paszke2019pytorch}
A.~Paszke, S.~Gross, F.~Massa, A.~Lerer, J.~Bradbury, G.~Chanan, T.~Killeen,
  Z.~Lin, N.~Gimelshein, L.~Antiga \emph{et~al.}, ``Pytorch: An imperative
  style, high-performance deep learning library,'' in \emph{Neural Information
  Processing Systems}, 2019.

\bibitem{chechik2009large}
G.~Chechik, V.~Sharma, U.~Shalit, and S.~Bengio, ``Large scale online learning
  of image similarity through ranking,'' in \emph{Iberian Conference on Pattern
  Recognition and Image Analysis}, 2009.

\bibitem{deng2018triplet}
C.~Deng, Z.~Chen, X.~Liu, X.~Gao, and D.~Tao, ``Triplet-based deep hashing
  network for cross-modal retrieval,'' \emph{IEEE Transactions on Image
  Processing}, vol.~27, no.~8, pp. 3893--3903, 2018.

\bibitem{Regmi_2019_ICCV}
K.~Regmi and M.~Shah, ``Bridging the domain gap for ground-to-aerial image
  matching,'' in \emph{IEEE International Conference on Computer Vision}, 2019.

\bibitem{Siam-FCANet}
S.~Cai, Y.~Guo, S.~Khan, J.~Hu, and G.~Wen, ``Ground-to-aerial image
  geo-localization with a hard exemplar reweighting triplet loss,'' in
  \emph{IEEE International Conference on Computer Vision}, 2019.

\end{thebibliography}
\end{document}